\documentclass[sigconf]{acmart}

\def\BibTeX{{\rm B\kern-.05em{\sc i\kern-.025em b}\kern-.08emT\kern-.1667em\lower.7ex\hbox{E}\kern-.125emX}}
    
\copyrightyear{2019}
\acmYear{2019}
\setcopyright{acmlicensed}
\acmConference[KDD '19]{KDD '19: ACM SIGKDD Conference on Knowledge Discovery and Data Mining}{August 04--08, 2019}{Anchorage, AK}
\acmBooktitle{KDD '19: ACM SIGKDD Conference on Knowledge Discovery and Data Mining, August 04--08, 2019, Anchorage, AK}
\acmPrice{15.00}
\acmDOI{10.1145/1122445.1122456}
\acmISBN{978-1-4503-9999-9/18/06}

\usepackage{algorithm}
\usepackage[noend]{algpseudocode}
\usepackage[flushleft]{threeparttable}
\usepackage{subcaption}

\usepackage{stfloats}

\newcommand\fixlen[2][10ex]{\def\fixwidth{#1}\fixlenpar#2\par\relax}
\long\def\fixlenpar#1\par#2\relax{%
  \fixlenword#1 \relax%
  \ifx\relax#2\relax\def\next{}\else\par\def\next{\fixlenpar#2\relax}\fi\next}
\def\fixlenword#1 #2\relax{%
  \makebox[\fixwidth][c]{#1}%
  \ifx\relax#2\relax\def\next{}\else\ \def\next{\fixlenword#2\relax}\fi\next}

\begin{document}

%
\title{Nostalgin: Extracting 3D City Models from Historical Image Data}

%

\author{Amol Kapoor}
\authornote{Both authors contributed equally to this research.}
\email{ajkapoor@google.com}
\affiliation{%
  \institution{Google Research}
  \city{New York}
  \state{NY}
}

\author{Hunter Larco}
\authornotemark[1]
\email{hunterlarco@google.com}
\affiliation{%
  \institution{Google Research}
  \city{New York}
  \state{NY}
}

\author{Raimondas Kiveris}
\email{rkiveris@google.com}
\affiliation{%
  \institution{Google Research}
  \city{New York}
  \state{NY}
}

%
\begin{abstract}
What did it feel like to walk through a city from the past?  
In this work, we describe Nostalgin (Nostalgia Engine), a method that can faithfully reconstruct cities from historical images. Unlike existing work in city reconstruction, we focus on the task of reconstructing 3D cities from historical images. Working with historical image data is substantially more difficult, as there are significantly fewer buildings available and the details of the camera parameters which captured the images are unknown. Nostalgin can generate a city model even if there is only a single image per facade, regardless of viewpoint or occlusions. To achieve this, our novel architecture combines image segmentation, rectification, and inpainting.
We motivate our design decisions with experimental analysis of individual components of our pipeline, and show that we can improve on baselines in both speed and visual realism. We demonstrate the efficacy of our pipeline by recreating two 1940s Manhattan city blocks. We aim to deploy Nostalgin as an open source platform where users can generate immersive historical experiences from their own photos.

\end{abstract}

%
%


%

%
\maketitle

\begin{figure}
    \centering
  \includegraphics[width=\linewidth]{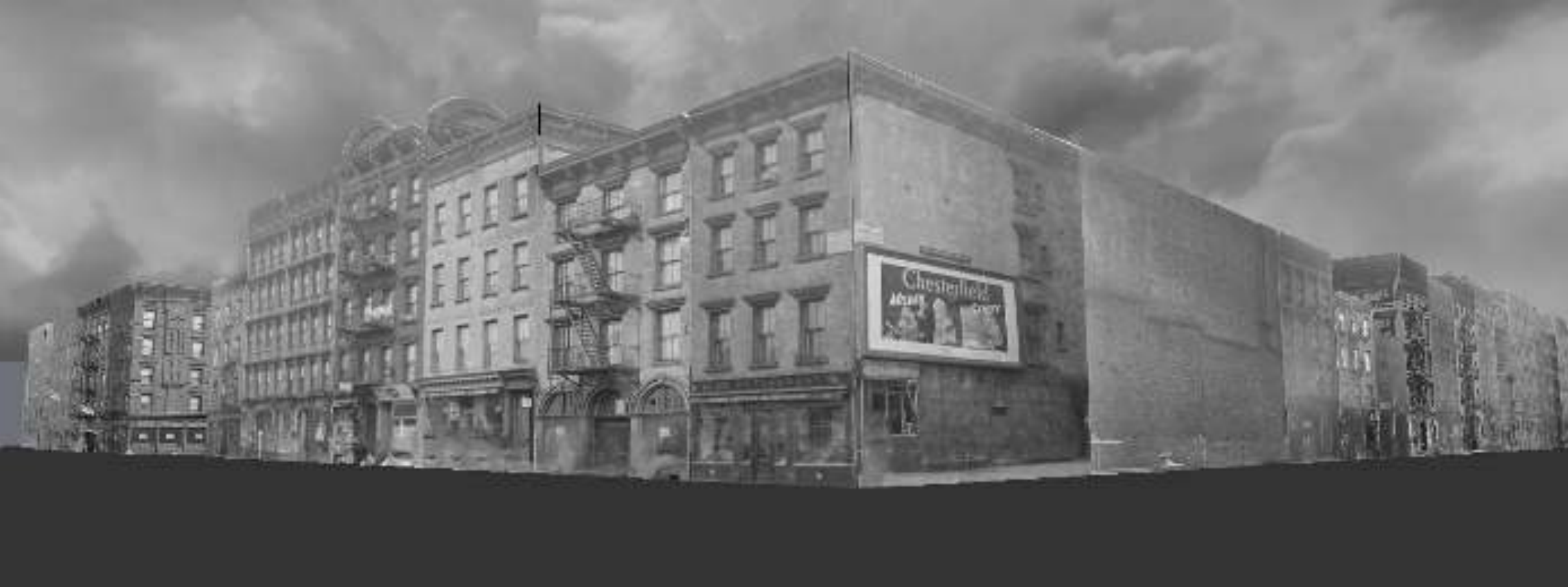}
  \caption{A 3D reconstruction of the NE Corner of 9th Avenue, 16th Street, New York, NY as it looked in the 1940s.}
  \Description{A 3D reconstruction of the NE Corner of 9th Avenue, 16th Street, New York, NY as it looked in the 1940s.}
  \label{fig:teaser}
\end{figure}

\section{Introduction}
There is significant interest in the automatic generation of 3D city models. Such models are used in Google Maps and Google Earth, in popular video games, in urban planning simulations, and more. However, these models are prohibitively expensive to create. Traditionally, large studios spend thousands of dollars and man-hours to create realistic worlds. Commercial procedural modeling engines are a powerful tool to address some of these issues, but they are limited in their accuracy and require significant manual effort to fine tune.

There is also significant interest in historical image data. Individuals are fascinated with historical data as a means of capturing nostalgia, pursuing education, connecting with family and elders, or preserving culture. Individuals are especially excited about historical data that allows them to interact with bygone eras, experiencing settings and environments that no longer exist. We note that city photography is a natural source of realistic detail, and that there is a significant wealth of historical and modern city imagery. With this in mind, we are interested in the problem of automatically generating city models from historical images of cities to expose historical data to users through an immersive walkthrough experience.

Historical images are difficult to access and even more so to utilize, especially in comparison to modern image and video data. Historical images are inherently more sparse than modern images, in that there are simply fewer available. Thus, when working with historical data, it is difficult to create large datasets with specific requirements, such as all images being occlusion-free, or taken from the same camera angle. It is also difficult to find multiple historical images of the same subject. Finally, historical metadata is nonexistent. Unlike modern images, which often come with EXIF information like geolocation and camera intrinsics, historical data often only includes raw pixel information.

Recent advances in computer vision have enabled the automatic recovery and extraction of missing information from images. Computers have gained the ability to semantically parse \cite{maskrcnn}, rectify \cite{zhangwhiteboard}, and inpaint \cite{yu2018free} images, and extract 3D scene understanding \cite{nishida2018procedural} from images. Research has been done to extract city geometries from images as well \cite{musialski2013survey}. Though these advances in computer vision are powerful, they often come with caveats and assumptions that make broad usage difficult. Many approaches require intrinsic or extrinsic camera parameters like focal length or relative geolocation; others are limited to toy datasets or require multiple input images; others still require fairly significant human intervention. These limitations carry over to 3D city generators. For example, the work of \cite{nishida2018procedural} requires a color image with few occlusions and a user-generated trace of the building model to create a single building. These limitations are not scalable and are unsuited for historical data, which is sparse and has few image guarantees.

In this work, we describe a scalable, modular 3D city generation pipeline named Nostalgin (Nostalgia Engine) that leverages, combines, and builds on advances in computer vision. Nostalgin is designed to uniquely handle the difficulties that arise when dealing with historical image data. Our novel contributions are as follows: 
\begin{enumerate}
    \item a combined deep and algorithmic approach to image segmentation that produces extremely tight segmentation masks;
    \item a novel approach to image rectification that can uniquely handle historical image data;
    \item a method for efficient deep image inpainting on extremely large images;
    \item a modeling system to place rectified facade images into a 3D world.
\end{enumerate}

For each section, we motivate our design choices and provide experimental analysis demonstrating the qualitative and quantitative efficacy of each component. We also analyze our overarching design and discuss approaches for better run-time and memory cost. Finally, we present two reconstructed blocks of Manhattan that are automatically generated using images taken from historical datasets of New York City in the 1940s.

\section{Related Work}
For non-deep-learning related work, we refer primarily to the  review in \cite{musialski2013survey}, which describes many important methods for accurate modeling 3D cities. These approaches can broadly be split by the type and amount of ingested data. Early approaches focus on street-level image data as an obvious source of information. Several works extract 3D geometry using multi-view image reconstruction \cite{agarwal2009building, agarwal2010bundle, irschara2007towards}, which often relies on understanding the general location of an image in order to make sense of the contents. These works contrast to single-view reconstruction, which use heuristics such as general shape and symmetry to mimic real world constructs \cite{jiang2009symmetric, kovsecka2002video, kovsecka2005extraction}, or are highly interactive and require user input \cite{horry1997tour, oh2001image, jiang2009symmetric}. More recently, development of hardware has made aerial imagery, satellite imagery, and LIDAR mapping significantly more viable. These forms of data allow for new kinds of 3D reconstruction. For example, \cite{kelly2017bigsur} proposes a method of combining street-level imagery, GIS footprints, and polygonal meshes (processed aerial images) to extract models, while \cite{korah2011strip} proposes utilizing aerial urban LIDAR scans. For completeness, we note that procedural modeling \cite{vanegas2010modelling} and manual modeling \cite{yin2009generating} are popular and well utilized in many practical applications. 

Due to the recent popularity of deep learning, a number of publications have proposed deep models to learn automatic reconstruction. Recent work has improved on extracting intrinsic camera parameters and object poses \cite{hara2017designing}, semantically parsing facades \cite{liu2017deepfacade}, or combining machine learning techniques with procedural grammars for reconstruction \cite{nishida2018procedural}. We note that many approaches to deep 3D reconstruction are not scalable and are very difficult to train outside of academic datasets (e.g. ShapeNet \cite{chang2015shapenet} which consists of low poly or voxel models that are not suitable for a city reconstruction task). 

In dealing primarily with historical data, we tackle a different task than many of the above methods. We cannot rely on any guarantees regarding multiple views, and do not have access to tools such as LIDAR, aerial data, satellites, or even cameras that measure parameters like focal length and position. We aim for a high degree of accuracy, potentially at the expense of detailed 3D features. Finally, we  desire a system that minimizes human input in order to generate entire cities at scale. 

\section{Proposed Method}
In this section, we describe the design of Nostalgin. We identify four key tasks in our image-to-model conversion process: image parsing, viewpoint normalization, occlusion removal, and 3D conversion. For each section, we provide a description of the sub-problem, our requirements for the solution, and the design of our final component. Experiments motivating our design choices are in Section \ref{sec:experiments}.

\begin{figure*}
	\centering
    \includegraphics[width=0.8\linewidth]{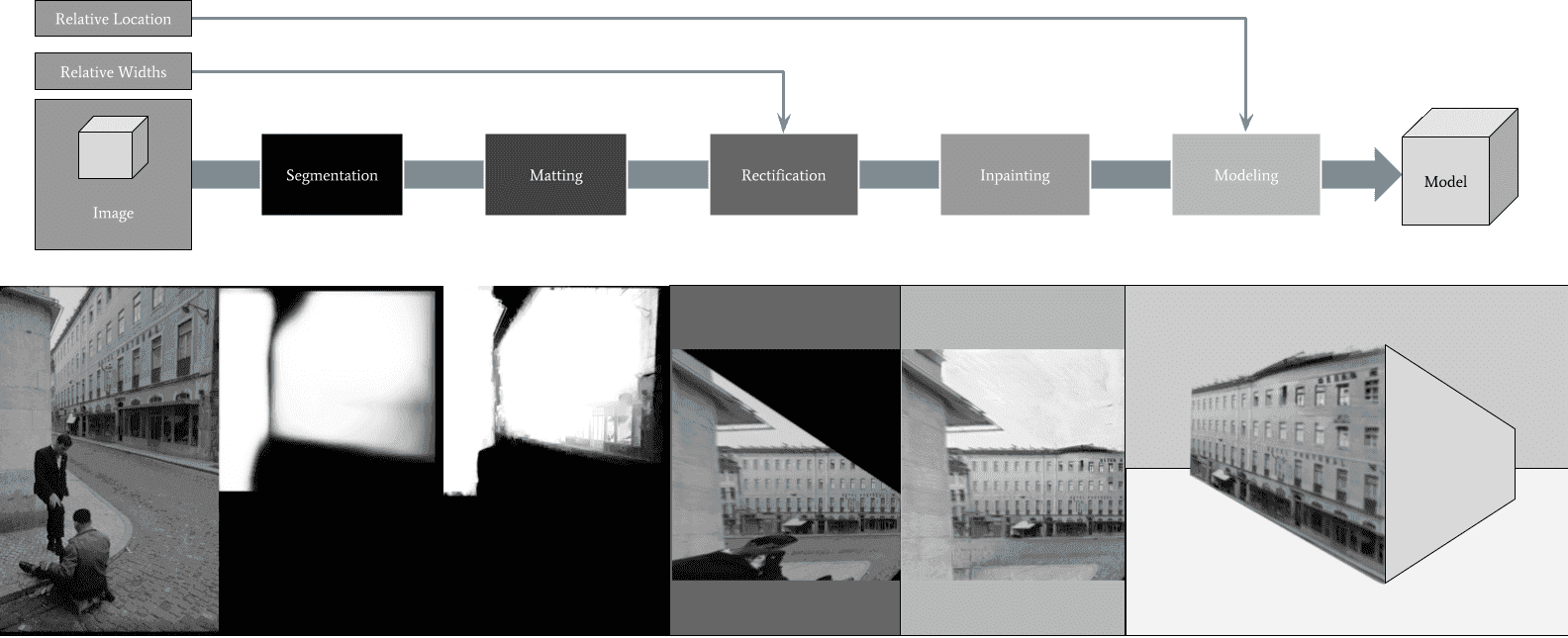}
    \caption{Components in our modeling pipeline.}
	\label{fig:components}
\end{figure*}

\subsection{Generalizations and Assumptions}

Because we are working with historical image data, we try to minimize the number of requirements related to the contents of the image and the metadata available. To that end, we design Nostalgin to be as general as possible without relying on anything other than the raw image data. At the same time, we purposely design our pipeline to require minimal human intervention so that it can work in massively distributed settings.

We generalize to the following conditions:
\begin{enumerate}
    \item as low as only one image per facade;
    \item possibly more than one facade in an image (see \ref{sec:segment});
    \item arbitrary aspect ratio and resolution;
    \item arbitrary viewing angle, and no apriori knowledge of viewing angle or relevant camera parameters (see \ref{sec:rect});  
    \item facade occlusions (see \ref{sec:inpaint});
    \item grayscale;
\end{enumerate}

These constraints significantly limit the amount of prior knowledge we can bring to bear in Nostalgin, making the underlying reconstruction task far more difficult and preventing usage of most prior work. However, these assumptions allow us to generalize to Nostalgin to real historical image data in a massively scalable way.

Our pipeline makes the following assumptions:
\begin{enumerate}
    \item the facades come from a Manhattan-world environment\footnote{A Manhattan-world assumption is the assumption that most buildings are relatively planar and lie on a cartesian grid, as in Manhattan. For example, we do not expect our pipeline to accurately handle domes.};
    \item images are weakly geotagged such that we are given the relative position of where each image was taken with respect to neighboring images (see \ref{sec:model});
    \item the width for each facade is known relative to other facades.
\end{enumerate}

\begin{figure}
	\centering
	\includegraphics[width=\linewidth]{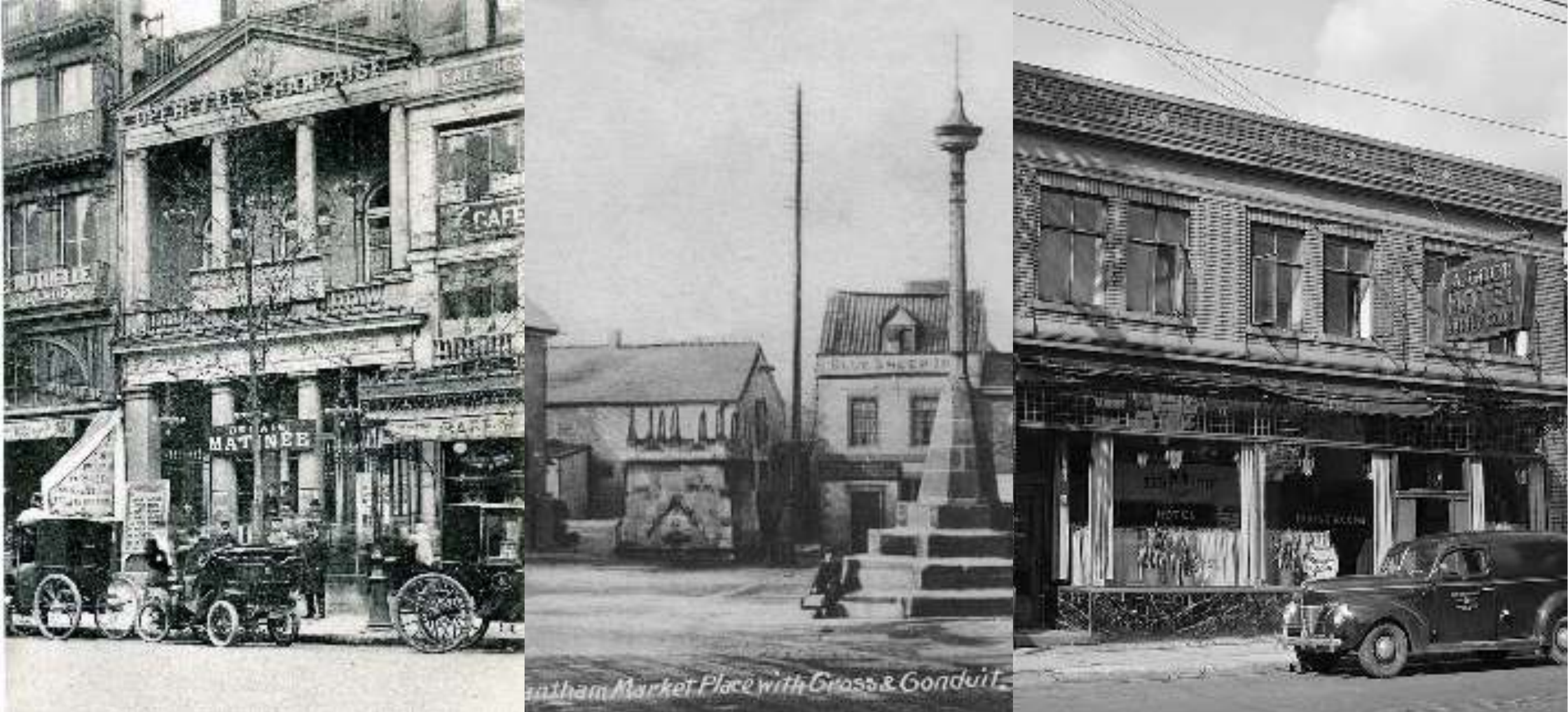}
    \caption{Examples of real world image data.}
    \Description{Examples of real world image data depicting images at various angles with occlusions.}
	\vspace{-0.5cm}
	\label{fig:real-world-images}
\end{figure}

\subsection{Image Parsing}
\label{sec:segment}
In order to gain insight from an image, we must first identify what objects are in the image and where they are actually located in pixel space. This includes identifying key objects of interest, such as one or many building facades, as well as identifying occlusions that may be blocking the full building image. Any kind of parsing should produce sharp boundaries around the parsed object in order to provide shape information to later components and to ensure that no pixel information is lost. 

Deep neural network models have made incredible strides in image segmentation and classification tasks. Thus, for our pipeline, we utilize the popular MaskRCNN deep neural network architecture \cite{maskrcnn}. We detect two classes of objects: building facades, and occlusions. In particular, we aim to label people and cars as occlusions. The MaskRCNN model is pretrained on COCO and fine tuned on a set of roughly 30k images that are manually labelled with masks around facades. A well known problem with this class of neural segmentation models is that the model struggles with providing extremely tight image boundaries. In order to address this issue, we add an image-gradient-based postprocessing step known as alpha matting \cite{laplacianmatting}. Alpha matting significantly improves the contours of our masks. Further analysis of the addition of alpha matting can be found in Section \ref{sec:seg-mat-impl}.

\subsection{Viewpoint Normalization}
\label{sec:rect}
The second task within our pipeline is to normalize the image with respect to camera viewpoint. This normalization takes the form of rectifying facades defined by a set of masks in an image. The goal of this normalization is to simplify downstream tasks to make it easier to extract depth and infer missing contextual information. Because we lack camera parameters and use real world images with many confounding objects in the scene, we develop our own rectification method based on previous work.

\subsubsection{Low-Signal Line Detection}
\label{sec:low-signal}
Almost all rectification approaches rely on accurate line detection in an image. Real world data often has complex structures that make line extraction difficult. Historical images additionally suffer from poor resolution, scanning artifacts, and image damage. As a result, we are unable to use off-the-shelf line detection methods such as the Probabilistic Hough Transform. Instead, we devise our own line detection algorithm that preserves lines that are good candidates for vanishing point detection and removes other lines. We provide brief analysis of other line detection methods in \ref{sec:linedet-impl}.

In order to capture as much signal as possible, we first run Canny edge detection with full connectivity and dynamically compute the thresholds given image median $\tilde{x}$ and hyperparameter $\lambda \in(0, 1)$ as follows.
\begin{equation}
l=\max(0, \tilde{x}(1-\lambda))
\quad\text{and}\quad
u=\min(255, \tilde{x}(1+\lambda))
\end{equation}
where $\lambda$ represents the tightness of our Canny thresholds. 
We then join all continuous points into contours.

To detect facade position, we want our line detector to only preserve straight lines. For each contour, we label every point as "linear" or "non-linear" by computing the second discrete derivative. Once labeled, non-linear points are removed and all remaining points are re-linked into new contours. We define the angle at a single point $p$ along the contour $C$ as
\begin{equation}\alpha(C, p)=arctan\left(\frac{\mathrm{d}p}{\mathrm{d}C}\right)\end{equation}
Using this, we define the left-hand second derivative as
\begin{equation}
\label{eq:lhd}
L_\alpha(C, p)=\frac{1}{k_s}\sum_{d=1}^{k_s}||\alpha(C, p)-\alpha(C, p-d)||_\theta
\end{equation}
and the right-hand second derivative as
\begin{equation}
\label{eq:rhd}
R_\alpha(C, p)=\frac{1}{k_s}\sum_{d=1}^{k_s}||\alpha(C, p)-\alpha(C, p+d)||_\theta
\end{equation}
where $||\theta_1-\theta_2||_\theta$ is the measure of the smallest angle between $\theta_1$ and $\theta_2$, and where $k_s$ is the window size of the discrete derivative.
Using Equations \ref{eq:lhd} and \ref{eq:rhd} and given a linearity threshold for the second derivative $t_\alpha$, we label each point along a contour and fit line segments to each locally-linear sub-contour using RANSAC \cite{ransac}. The algorithm for defining local linearity is provided in the Appendix.

\subsubsection{Vanishing Point Detection}

Vanishing point detection helps convert lines into depth information. Detecting vanishing points is a task traditionally solved in two steps: first, detected line segments are used to accumulate a list of potential vanishing point candidates; and second, the segments are used to rank the candidates. 

\begin{figure}
  \begin{subfigure}{0.48\linewidth}
    \includegraphics[width=\linewidth]{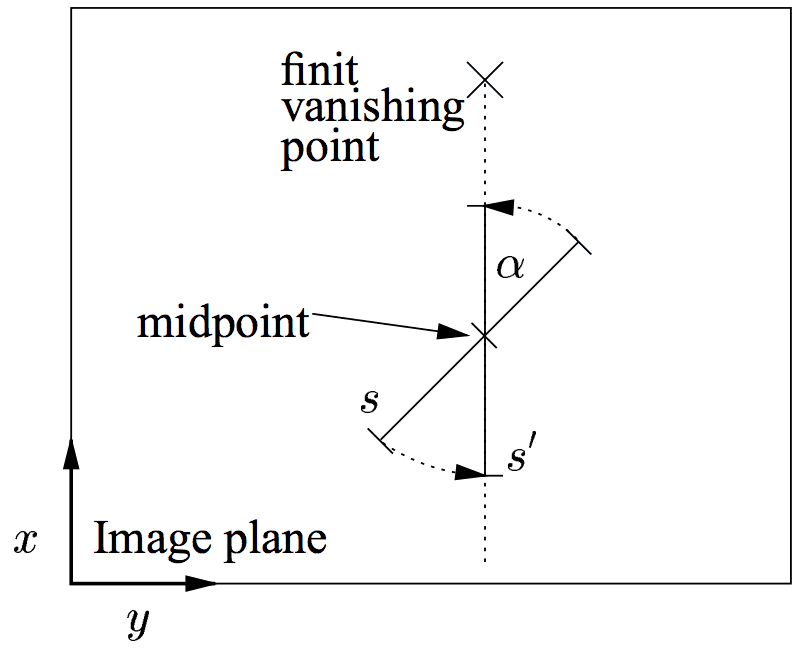}
    \caption{Explanation for the distance function $d(vp, s)=\alpha$ between a line segment $s$ and a finite vanishing point $vp$.}
    \Description{Explanation for the distance function $d(vp, s)=\alpha$ between a line segment $s$ and a finite vanishing point $vp$.}
  \end{subfigure}
  \hfill
  \begin{subfigure}{0.48\linewidth}
    \includegraphics[width=\linewidth]{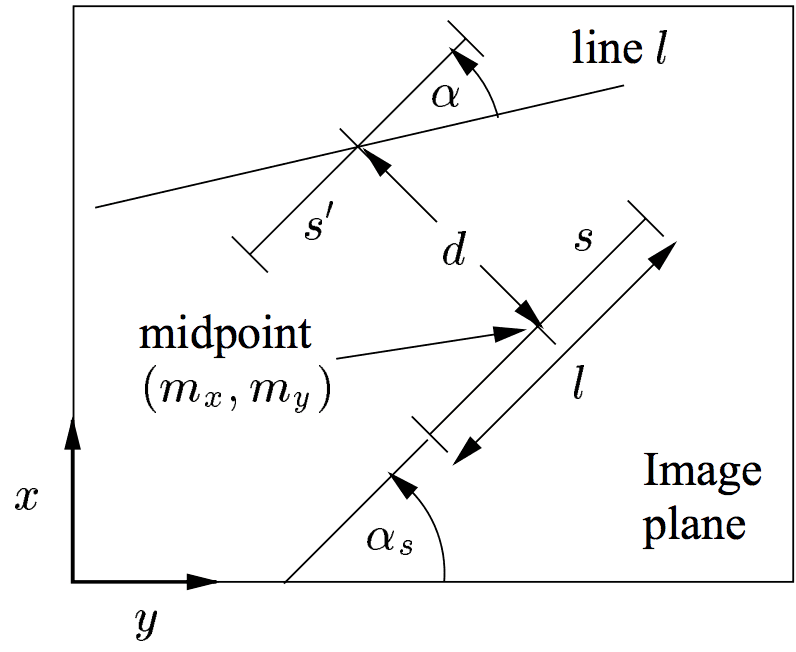}
    \caption{Explanation for the distance function $d(l, s)$ between infinite vanishing point $l$ and segment $s$.}
    \Description{Explanation for the distance function $d(l, s)$ between infinite vanishing point $l$ and segment $s$.}
  \end{subfigure}
  \caption{Duplicated from \cite{rothervps}: distance functions used to determine fit between a line segment and an (in)finite vanishing point.}
  \Description{Duplicated from \cite{rothervps}: distance functions used to determine fit between a line segment and an infinite or finite vanishing point.}
  \label{fig:rother-distance}
\end{figure}

During accumulation, we reduce the candidate search space by deduplicating collinear line segments and vanishing point candidates using quantization within our error bounds (see \ref{sec:spacered}). During voting, we modify Rother's voting function such that the resulting weights correspond to the percent of evidence accounted for by the vanishing point\footnote{For example, a value of 1.0 indicates a perfect match with all segments, whereas a value of 0.5 indicates that roughly half of segments match.}. Thus, we define that for a candidate vanishing point $a$, set of line segments $S$, facade mask $m$, and alignment threshold $t_a$
\begin{equation}vote(a, S, m)=\frac{\sum_{s}^{S}\left[||s||_2\omega(s, m)(1-\frac{d(a, s)}{t_a})\right]}{\sum_{s}^{S}\left[||s||_2\omega(s, m)\right]}\end{equation}
where $\omega(s, m)$ is a weighting function defined as the count of pixels on segment $s$ within mask $m$ normalized by the length of segment $s$. This is done to ensure that only pixels within a facade mask vote towards vanishing points for that facade. Note that function $d(a, s)$, shown in Figure \ref{fig:rother-distance}, is taken directly from \cite{rothervps}. Also, note that the alignment threshold $t_a$ measures the maximum distance $d(a, s)$ between a vanishing point and line segment that still constitutes alignment.

We select the most highly weighted vanishing point as scored using the corresponding facade mask $m$, and the second most highly weighted vanishing point that is at least $t_o$ degrees offset from the first, to find two vanishing points representing facade orthogonal lines.

\subsubsection{Quadrangle Estimation}
\label{sec:quadrangle-estimation}

\begin{figure}
  \begin{subfigure}{0.48\linewidth}
    \includegraphics[width=\linewidth]{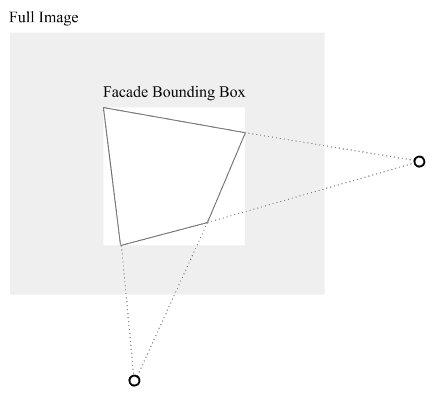}
    \caption{Vanishing points and facade.}
    \Description{The vanishing points and corresponding facade.}
  \end{subfigure}
  \hfill
  \begin{subfigure}{0.48\linewidth}
    \includegraphics[width=\linewidth]{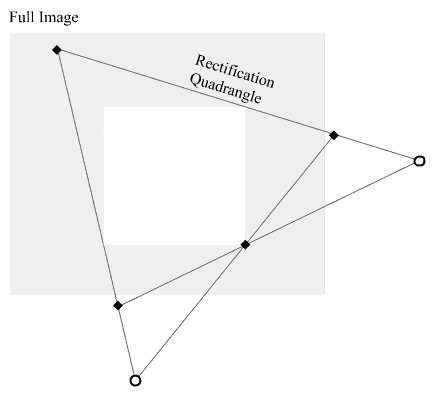}
    \caption{The rectification quadrangle.}
    \Description{The rectification quadrangle.}
  \end{subfigure}
  \caption{Computation of the facade bounding-quadrangle.}
  \Description{Computation of the facade bounding-quadrangle.}
  \vspace{-0.5cm}
  \label{fig:rectification-quadrangle}
\end{figure}

Once two vanishing points have been chosen, forming a vanishing point aware minimum-bounding quadrangle -- i.e. the smallest quadrangle that adheres to the facade's two vanishing points and also includes all of the facade's masked pixels -- is relatively straight-forward. Given the facade's pixel-mask, we can readily compute the bounding box for the facade. From this, we project lines from each vanishing point to the nearest corners of the bounding box and form a quadrangle from the four intersections created (see Figure \ref{fig:rectification-quadrangle}). The resulting quadrangle is a representation of the facade-plane projected onto the image-plane and resized to contain all masked facade-pixels.

\subsubsection{Rectification}
\label{sec:aspect}

In order to finish the rectification of the facade in the image, we need to predict the aspect ratio of the final image. Many existing approaches are able to leverage known camera parameters; however, working with general historical data naturally precludes any reliance on such information. Instead, we predict that the camera's principal point is the center of the image and that there is no skew in the image. This requires us to estimate only the focal length, which can be approximated using vanishing point geometry. We note that this can result in some error, but qualitative results suggest that the visual impact of imperfect focal length prediction is minimal within certain reasonable error bounds.

With a single quadrangle per-facade and an approximate focal distance, we can directly apply \cite{zhangwhiteboard} to determine the aspect ratio of the resulting rectified image. The aspect ratio and quadrangle vertices together give four corresponding points between the facade-plane and the rectification-plane which are sufficient to compute a rectification homography. This allows us to manipulate the facade in the image such that it looks like the camera pose has shifted to the front of the facade. Once we have the rectified facade and the appropriate aspect ratio, we use the given width of the facade image to scale the facade relative to its real world location context.

\subsection{Occlusion Removal}
\label{sec:inpaint}
The third task is to normalize the image with respect to occlusions. This component also ingests a set of masks and an image, and outputs an inpainted image with the inpainting occurring in the masked locations. Importantly, any approach used for inpainting had to handle fairly high-resolution images (larger than 800 x 800 pixels) and had to work for large, arbitrary masks. 

\subsubsection{Inpainting Methods}

We examine several deep and algorithmic approaches, and describe our analysis in Section \ref{sec:inpaint-impl}. We build on the Free Form inpainter suggested by \cite{yu2018free}. Specifically, we create a two-stage conditional fully-convolutional GAN with an SNPatch Discriminator, trained on black and white images. 

The Free Form approach is memory intensive for images larger than 250x250 pixels. In order to solve this issue, we decrease the width of the model by nearly 25\% and double the stride of the contextual attention layer. We also develop a `Low Memory' Inpainter that takes an input image and a set of masks, splits the image to only include the mask and a small surrounding area based on a preset context radius, and inpaints each split separately. These approaches to decreasing memory usage also decrease accuracy. We discuss these tradeoffs in \ref{sec:inpaint-impl-scale}.

\subsubsection{Dataset}

A convenient aspect of the Free Form method is that it learns how content extends across 2D geometry instead of learning to represent a specific object class. This is especially important in historical settings, where it is difficult to collect a large dataset of a specific object. We collect a dataset of 10M modern and historical images. We require only that each image has at least one facade in the image. We convert each image to black and white, and train the model on 600x400px random crops. We refer to this dataset as the 10M dataset. 

\subsection{Modeling}
\label{sec:model}

The final task is to generate a 3D city model. This component expects a set of cropped facade images that are to-scale, each with relative location information. For this work, we assume that the buildings will appear in a grid-like city block formation; as such, the model only requires the left- or right-side neighbors of each facade, whether two facades come from the same building, and the location of each block relative to each other.  

We utilize the facade location data to create a chain of facades that wrap around each block, and then place the blocks relative to each other. For each facade, we create a cuboid 3D model with matching proportions; if more than one facade is given for the same building, we can exactly specify the parameters of the cuboid model. We provide the algorithm for placing buildings within a block in the Appendix. The complete algorithm for placing blocks is a trivial extension.

For each cuboid model, we apply the relevant input facade images as textures. If four facades are not given, we tile the given facades around all four sides of the cuboid. We make all parts of the image that are not part of the facade transparent before texture application, utilizing matting masks to determine where facade boundaries are.

\section{Experiments}
\label{sec:experiments}

In this section we motivate specific design decisions through qualitative and quantitative measures of performance for each subcomponent in the larger 3D modeling pipeline. For all experiments, see the Appendix for additional details on hyperparameter settings, evaluation datasets, loss calculations, and more.

\subsection{Segmentation and Matting}
\label{sec:seg-mat-impl}

For image segmentation, we utilize a MaskRCNN architecture. MaskRCNN is one of the most popular image segmentation architectures due to its ease of implementation and effectiveness in applied settings. We train the MaskRCNN model to select facades and occlusions (people, cars) in images\footnote{We note that it is easy to train for more classes of occlusions; for this proof of concept work, we selected the two most common occlusion types.}. However, we find that MaskRCNN masks degrade close to segmentation boundaries. This results in significant decrease of quality in later parts of the pipeline.

\begin{figure}
	\centering
	\includegraphics[width=\linewidth]{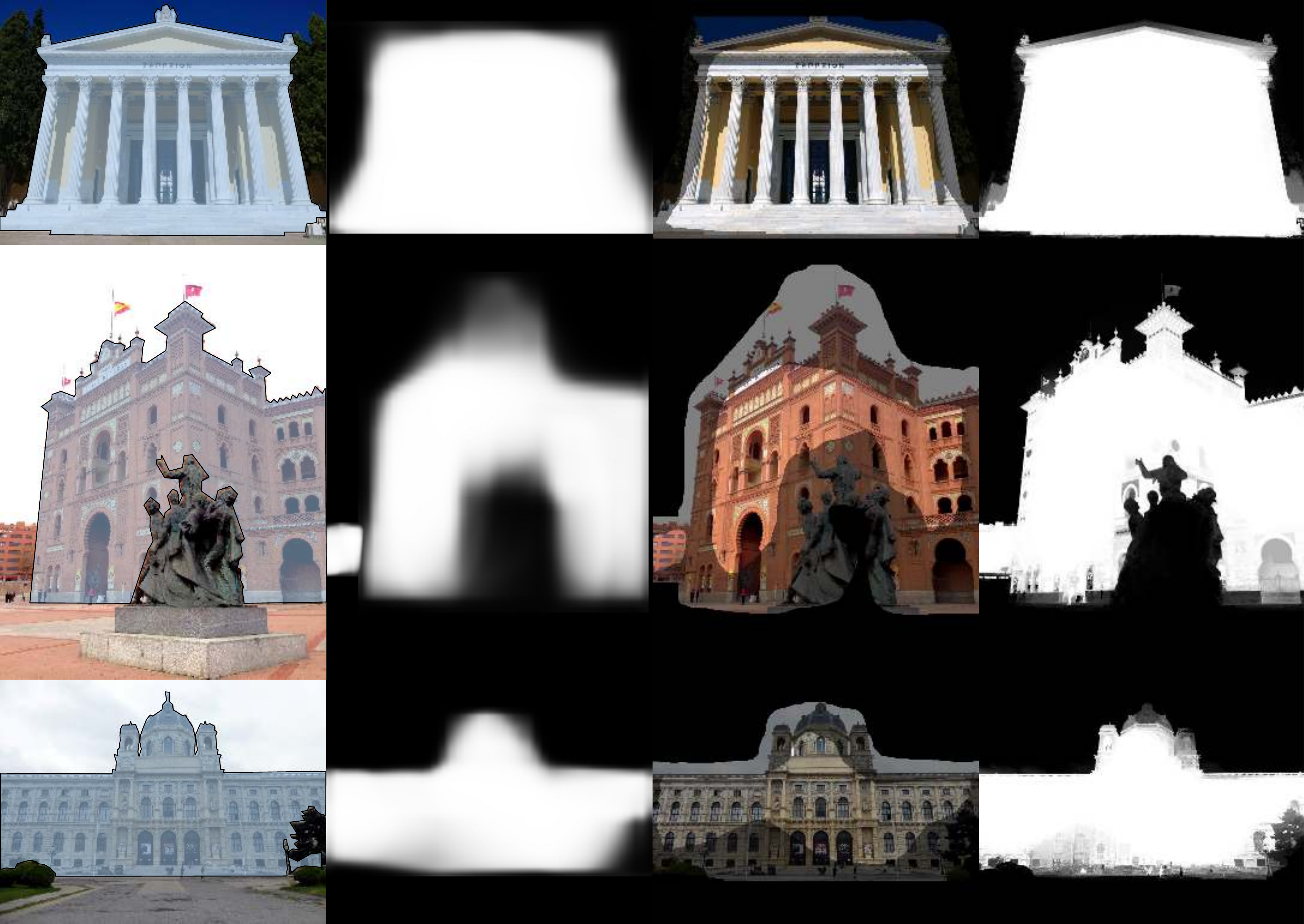}
	\caption{Qualitative analysis of matting improvements to segmentation. From left to right, we show the input image with the manually labeled ground truth, the MaskRCNN output, the generated trimap, and the output of alpha matting. Best viewed with zoom.}
	\Description{Qualitative analysis of segmentation and matting.}
	\vspace{-0.5cm}
	\label{fig:segment-qual}
\end{figure}

In order to produce tighter image boundaries, we examine image matting algorithms. We convert the output MaskRCNN model to a trimap, using the probabilities of the MaskRCNN to map the range of 5\% to 95\% as uncertain. We then apply the image gradient-based alpha matting algorithm from \cite{laplacianmatting}.  Using manually labeled ground truth masks, we compare precision, recall, $l1$ loss, and $l2$ loss in Table \ref{tab:matting}. We show qualitative results in Figure \ref{fig:segment-qual}. 

Matting increases recall by 8\%, and decreases precision by 11\%. We prefer a high recall model because later components of Nostalgin rely on line data, and so pulling out more lines for a facade is empirically beneficial. However, on manual inspection of the qualitative results, we find that the masks produced by matting capture boundaries \textit{better} than the manually labeled ground truth around difficult edges that manual labelling ignored. This explains a significant amount of the error in both precision and recall. 

Masks produced by matting had slightly improved loss on both $l1$ and $l2$ metrics. Manual inspection of qualitative results showed that almost all of this improvement was localized around object edges. Matting masks showed stronger weighting along actual boundaries of each segmented object. By contrast, MaskRCNN masks had a `fade' effect along object edges, resulting in low probability weights being given to the strongest directional lines in the captured object.

Finally, we note the high variance among all measured metrics. We attribute this to the inherent variation in our ground truth data:  because we did not explicitly capture every facade in every image, images where the MaskRCNN missed a facade or captured one that was not in ground truth caused huge variations in these metrics. 

\begin{table}
  \begin{threeparttable}
    \caption{Quantitative Segmentation Comparison}
    \label{tab:matting}
    \begin{tabular}{lcccc}
      \toprule
      & Precision & Recall & $l1$ & $l2$\\
      \midrule
      MaskRCNN & 86.1 $\pm$ 10.3 & 78.2 $\pm$ 5.3 & 10.9 $\pm$ 8.1 & 8.2 $\pm$ 7.9 \\
      Matting & 75.2 $\pm$ 13.1 & 86.5 $\pm$ 8.2 & 10.4 $\pm$ 8.5 &  7.6 $\pm$ 7.3\\
      \bottomrule
    \end{tabular}
  \end{threeparttable}
\end{table}

\subsection{Rectification}
\label{sec:rect-impl}

\subsubsection{Analysis of Line Detection Methods}
\label{sec:linedet-impl}

Traditional approaches for line detection such as Probabilistic Hough Transform or LSD \cite{lsd} are attractive because they require no hyperparameter tuning and can be applied with little-to-no development cost using tools such as OpenCV. However, we observe that these line detectors yield poor rectifications, as occlusions such as people, tree, and cars in addition to building ornamentation such as domes, arches, and statues dilute the signal of the facade. Specifically, off-the-shelf line detectors end up accommodating `curvy' occlusions by segmenting each contour into countless little lines at varying angles.

\begin{figure}
\begin{subfigure}{0.32\linewidth}
\includegraphics[width=\linewidth]{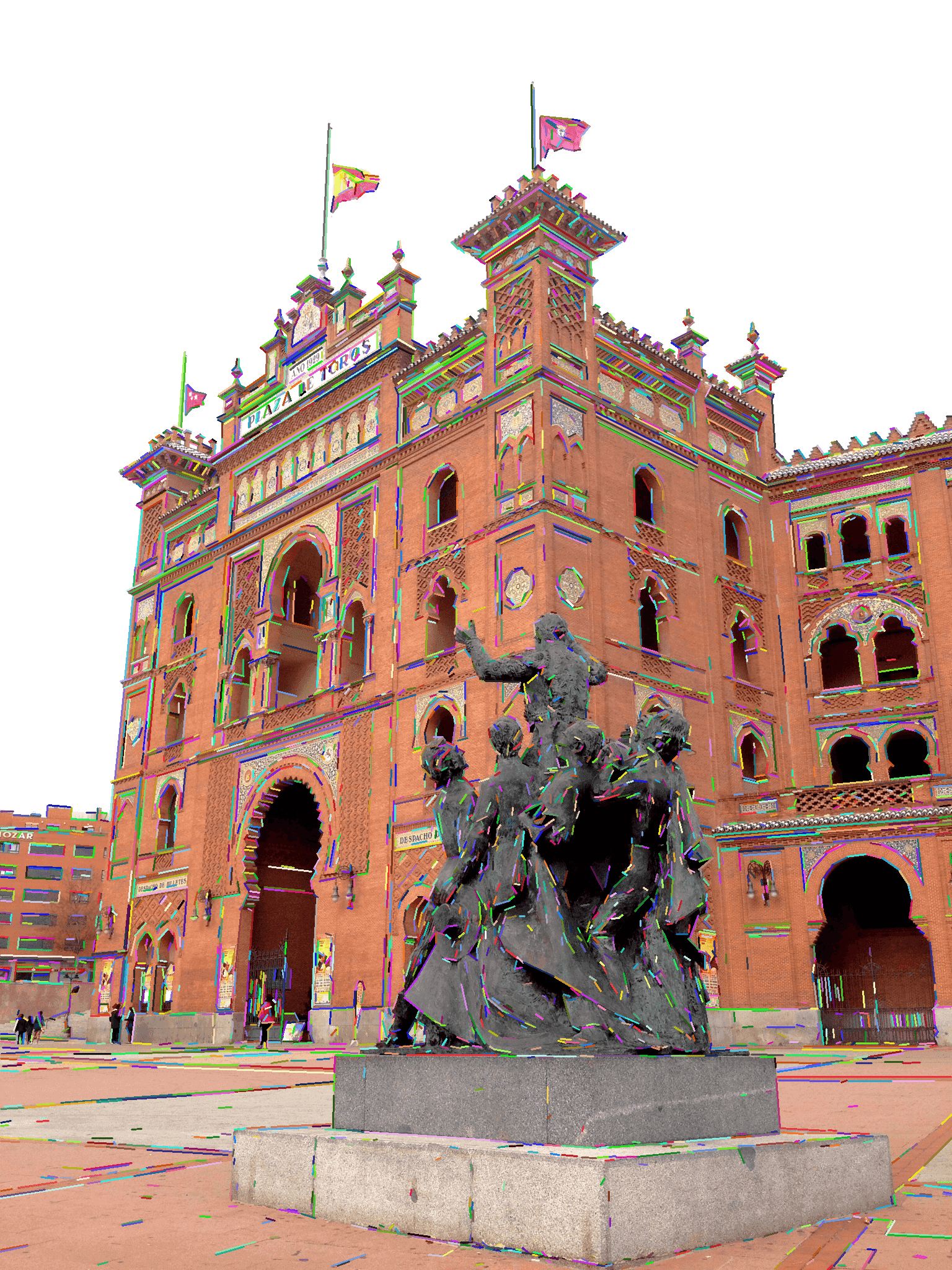}
\caption{LSD}
\end{subfigure}
\hfill
\begin{subfigure}{0.32\linewidth}
\includegraphics[width=\linewidth]{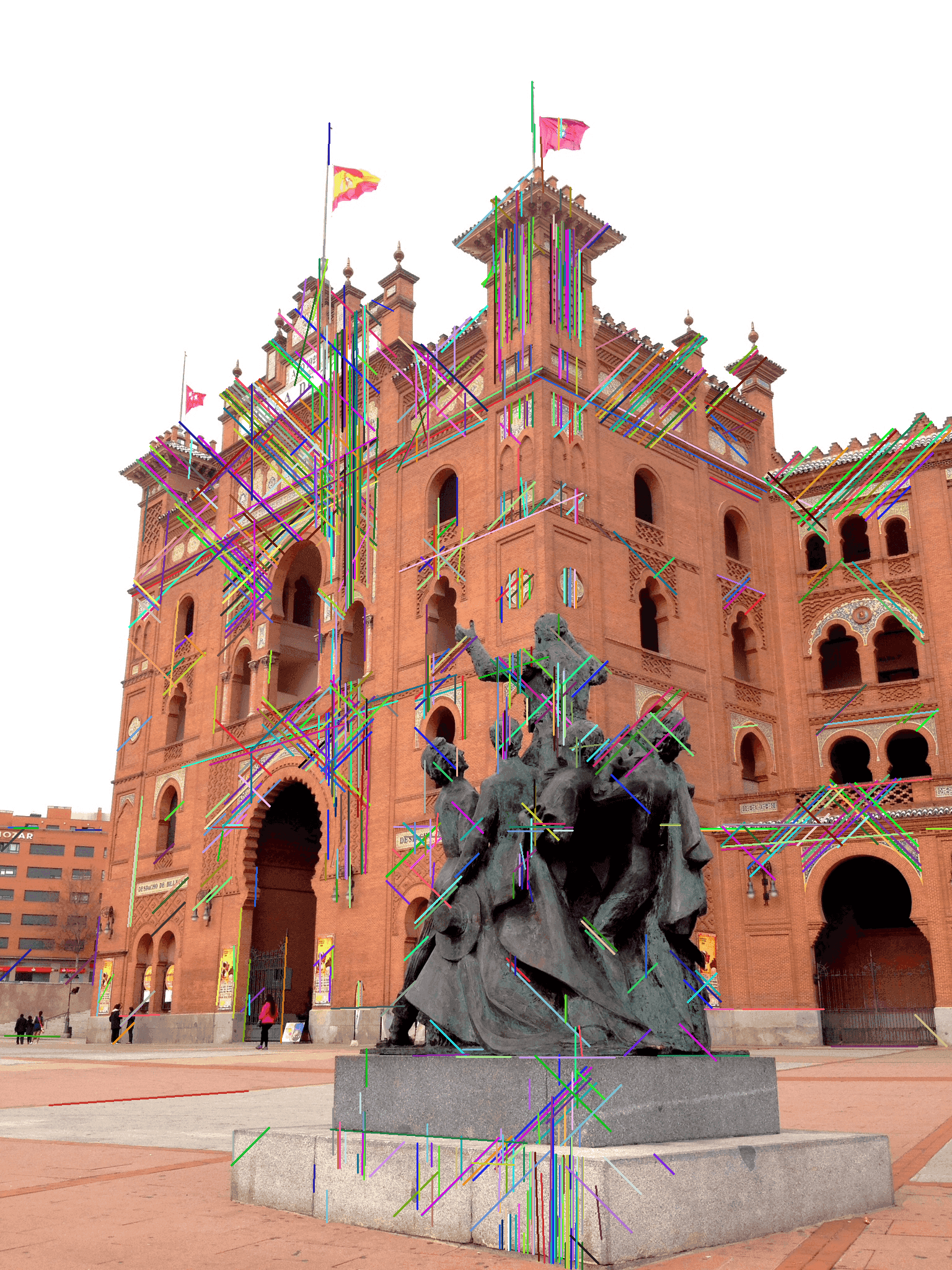}
\caption{Hough Transform}
\end{subfigure}
\hfill
\begin{subfigure}{0.32\linewidth}
\includegraphics[width=\linewidth]{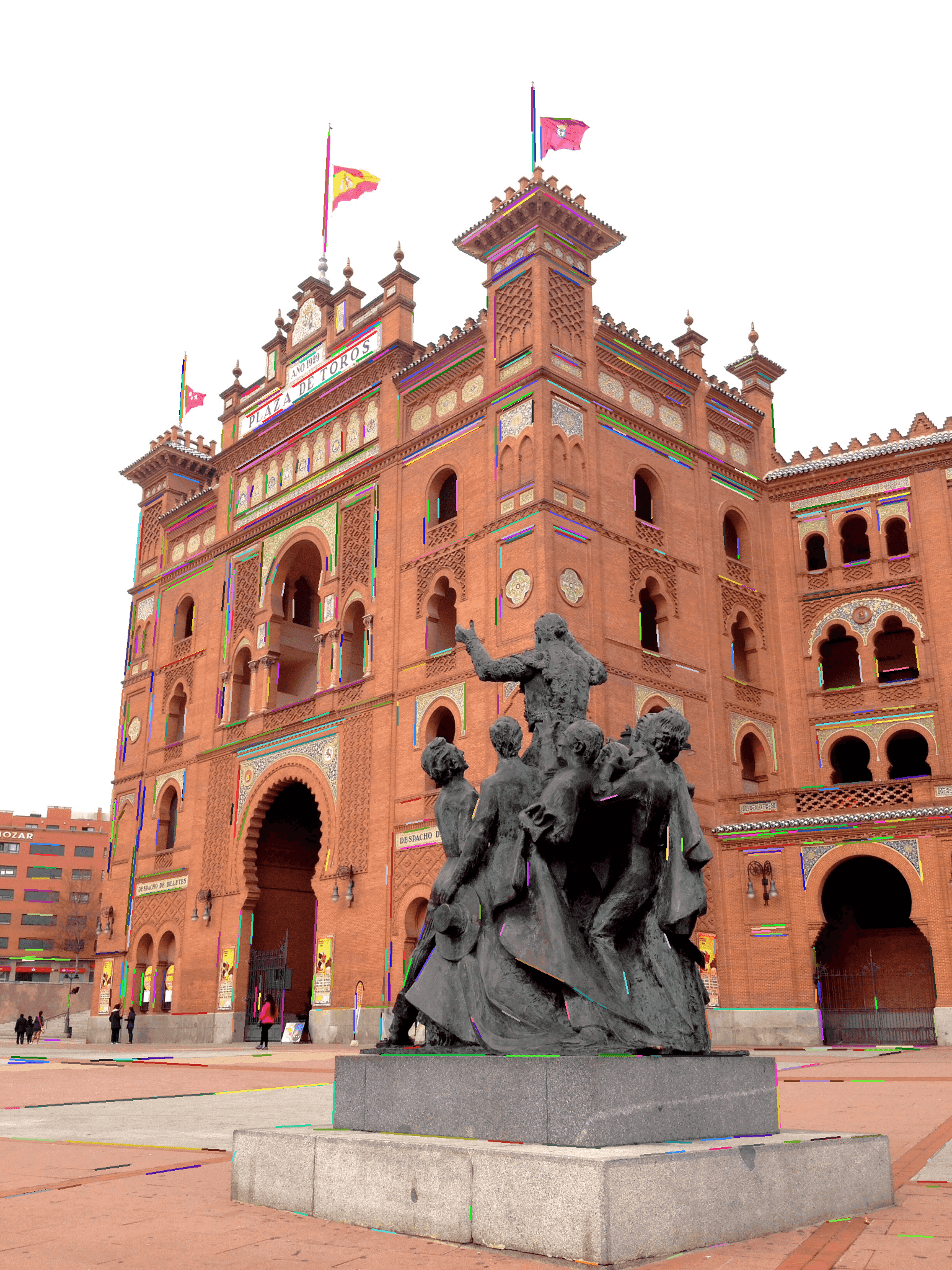}
\caption{Nostalgin}
\end{subfigure}
\hfill
\caption{Qualitative analysis of line-detection methods. Best viewed with zoom.}
\Description{Qualitative analysis of line-detection methods.}
\label{fig:line-detection-methods}
\vspace{-0.5cm}
\end{figure}

Our proposed method strengthens the signal of the facades and removes line data coming from ornamentation and occlusions. See Figure \ref{fig:line-detection-methods} for a qualitative comparison of the proposed methods and traditional approaches. We note that ornamentation along the roof and the occluding statue are less represented when using our proposed method. This allows our pipeline to focus on lines that actually provide depth information about the plane of the facade.

\subsubsection{Vanishing Point Space Reduction}
\label{sec:spacered}

In our initial rectification implementation, we noticed that the process of selecting, accumulating, and voting on appropriate vanishing points was responsible for over half of our run-time. We recognized that many of the vanishing points that were being analyzed were duplicates or near duplicates. We took efforts to decrease the vanishing point analysis space by reducing colinear line segments and infinite vanishing point segments. We measure the percent reduction of the search space and the wall time in Table \ref{tab:globalspacereduction}. Combining these two deduplication processes reduces our total global search space by 44\%, which in turn allows us to decrease the wall clock time used to calculate vanishing points by 35\%.

\begin{table}
  \begin{threeparttable}
    \caption{\% Reduction of Vanishing Point Candidates}
    \label{tab:globalspacereduction}
    \begin{tabular}{lcc}
      \toprule
      & Search Space & Wall Time \\
      \midrule
      Deduplicate Collinear Segments & 41.0 $\pm$ 16.1 & 33.1 $\pm$ 29.8 \\
      Deduplicate Infinite VPs & 11.4 $\pm$ 19.1 & 10.1 $\pm$ 7.5 \\
      \textit{Combined} & \textbf{44.3 $\pm$ 18.7} & \textbf{35.6 $\pm$ 37.7} \\
      \bottomrule
    \end{tabular}
  \end{threeparttable}
\end{table}

\subsection{Inpainting}
\label{sec:inpaint-impl}

\subsubsection{Analysis of Inpainting Methods}

We examine several traditional and deep learning approaches to inpainting. 
As far as we are aware, there are no industry standard methods of quantifying the quality of an inpainted image. In this work, we follow \cite{yu2018free} and use mean $l1$ and $l2$ loss as quantitative metrics. We note that these metrics have tenuous relation to the visual outcome of inpainting, especially when the inpainter is purposely attempting to remove an object or objects from a scene; thus, we rely heavily on qualitative results.

Traditional approaches to inpainting are promising because they require minimum or no training time and can handle large images with relatively small increases in memory cost (though often with a very large increase in computation time). Such approaches rely on local similarity metrics that allow semi-accurate `copy paste' operations. Diffusion based methods such as the Navier Stokes method \cite{bertalmio2001navier} propagate immediate neighboring pixel information based on image gradient information; while patch based methods such as PatchMatch \cite{barnes2009patchmatch} extend groups of local pixels based on low level features. These methods are powerful, but scale poorly to larger masks both in terms of quality and run-time.

In contrast, deep approaches to inpainting are promising because they learn semantic features across an entire image. Further, the run-time for deep approaches is often not a function of mask size. Several deep approaches, such as Semantic Inpainting \cite{yeh2017semantic}, are not resolution independent. These models require train and inference image sizes to be the same due to the presence of non-convolutional layers in the model. Other deep approaches such as Inpainting with Contextual Attention \cite{yu2018generative} are dependent on specific a mask shape and location and do not generalize well to arbitrary masks. 

\begin{figure}
	\centering
	\begin{subfigure}{\linewidth}
	    \centering
        \includegraphics[width=0.432\linewidth]{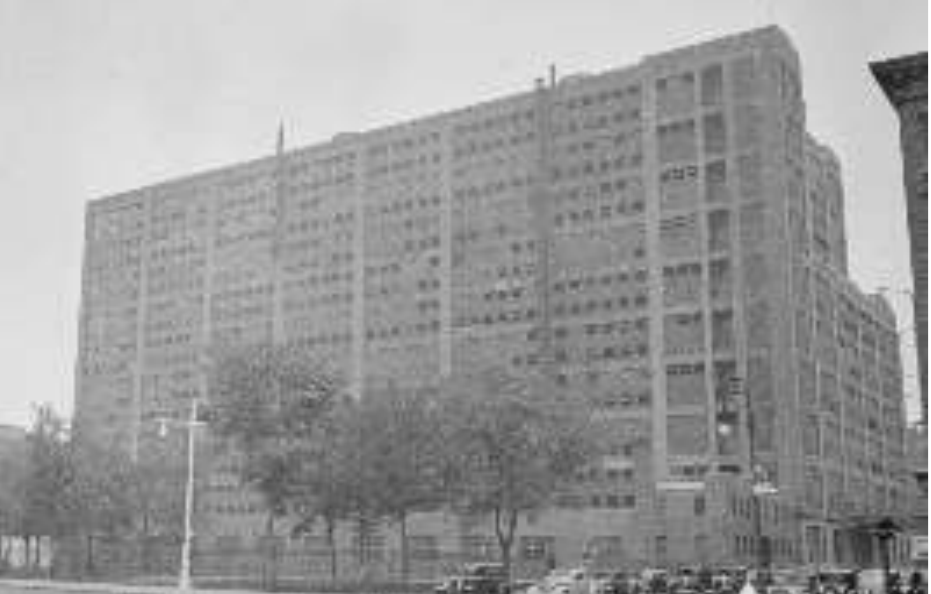}
        \includegraphics[width=0.432\linewidth]{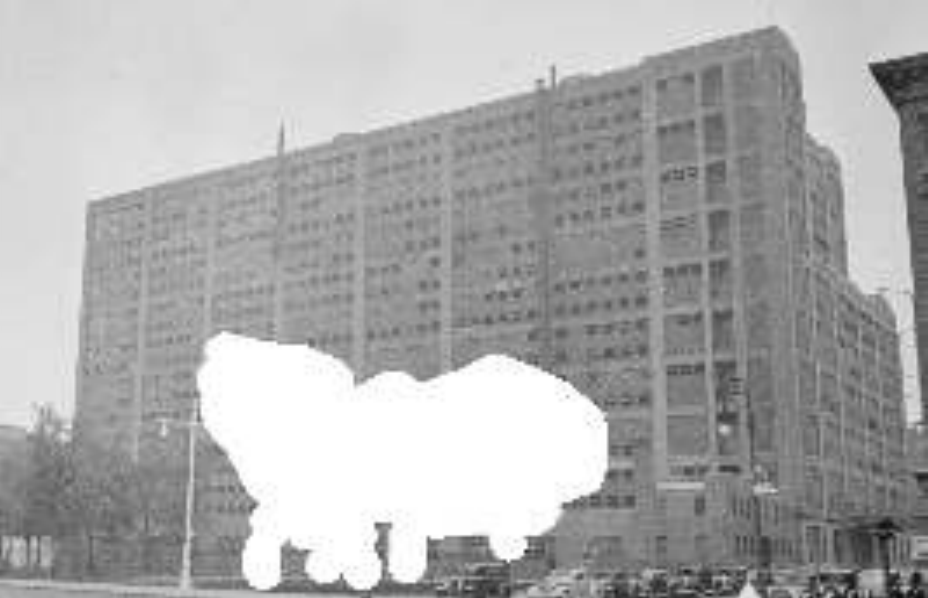}
        \caption{Original Image and Input Mask}
    \end{subfigure}

    \begin{subfigure}{0.432\linewidth}
        \includegraphics[width=\linewidth]{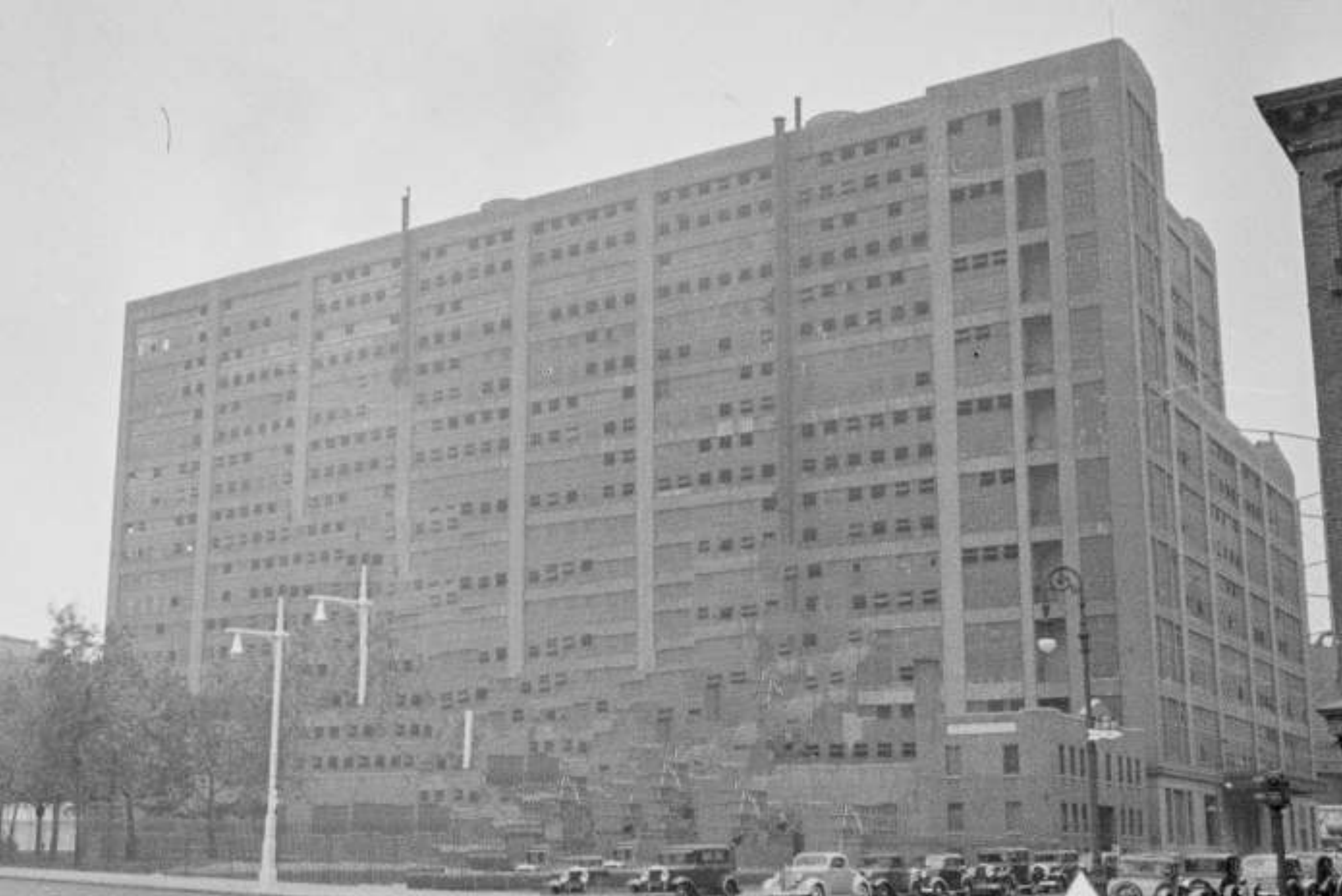}
        \caption{PatchMatch}
    \end{subfigure}
    \begin{subfigure}{0.432\linewidth}
        \includegraphics[width=\linewidth]{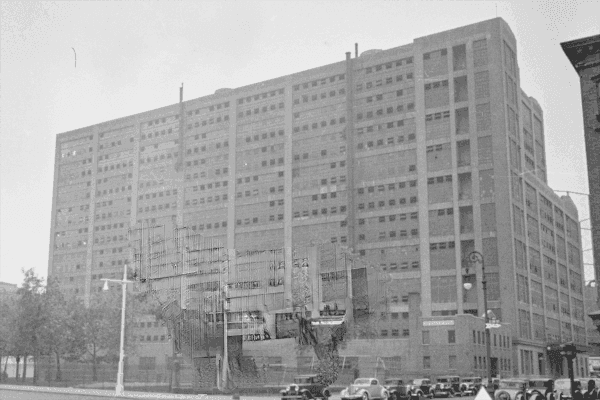}
        \caption{Free Form}
    \end{subfigure}
    
    \begin{subfigure}{0.432\linewidth}
        \includegraphics[width=\linewidth]{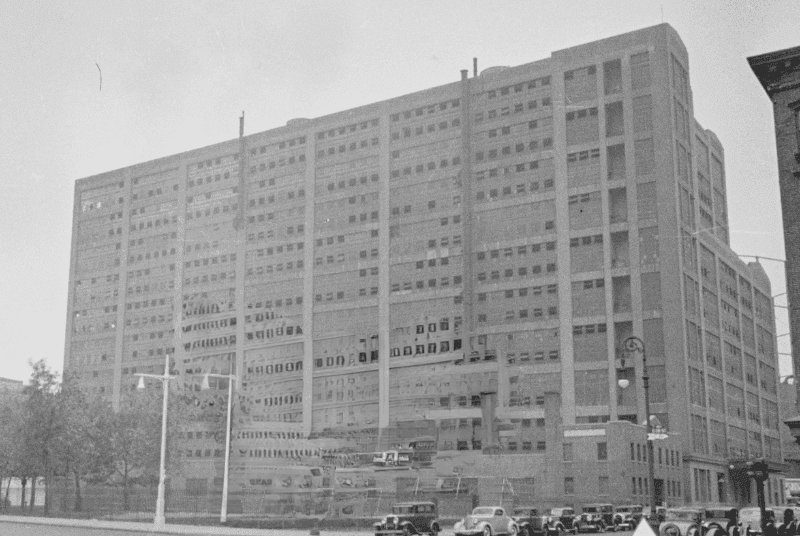}
        \caption{Nostalgin}
    \end{subfigure}
    \begin{subfigure}{0.432\linewidth}
        \includegraphics[width=\linewidth]{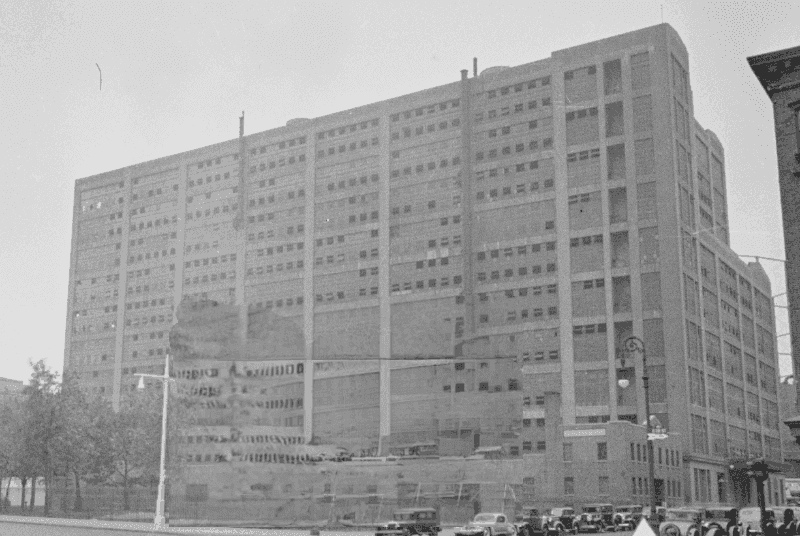}
        \caption{Nostalgin (Low Memory)}
    \end{subfigure}
	\caption{Qualitative analysis of inpainting methods. Best viewed with zoom. Image courtesy of the New York Municipal Archives.}
	\Description{Qualitative analysis of inpainting methods.}
	\vspace{-0.5cm}
	\label{fig:inpaint-analysis}
\end{figure}

The Free Form method proposed in \cite{yu2018free} fulfills our requirements, and we adapt it for this work. We discuss methods to improve the scalability of this approach in \ref{sec:inpaint-impl-scale}; we decide to decrease model capacity in exchange for better run-time. In Figure \ref{fig:inpaint-analysis} and Table \ref{tab:inpainter-quant} we respectively provide qualitative and quantitative analysis of several of the mentioned methods. 

Both versions of our model perform better than other methods on the $l1$ metric, besides the baseline Free Form model. We perform slightly worse on the $l2$ metric, indicating higher variability in our output. We note that these results are expected; because our model is 25\% slimmer and uses a larger stride, it has a smaller model capacity. We further expect some performance degradation because our models were trained on 10M and evaluated on Places2. However, in our qualitative measures, we observe little difference between our methods and the Free Form approach; in some cases, trained models with higher $l1$ and $l2$ losses performed `better' in terms of visual appeal and realism.

\begin{table}
  \begin{threeparttable}
    \caption{Inpainting Quantitative Comparison}
    \label{tab:inpainter-quant}
    \begin{tabular}{lcccc}
        \toprule
        & Per Pixel $l_1$ Loss & Per Pixel $l_2$ Loss \\
        \midrule
        PatchMatch* & 11.3 & 2.4 \\
        Global\&Local* & 21.6 & 7.1 \\
        ContextAttention* & 17.2 & 4.7 \\
        PartialConv* & 10.4 & 1.9 \\
        FreeForm* & 9.1 & 1.6 \\
        \midrule
        Nostalgin & 9.8 $\pm$ 4.2 & 2.5 $\pm$ 4.2 \\
        Nostalgin (Low Memory) & 10.4 $\pm$ 4.1 & 2.8 $\pm$ 1.8 \\
        \bottomrule
    \end{tabular}
    \begin{center}
    \textit{Loss values for starred methods taken from \cite{yu2018free}.}
    \end{center}
  \end{threeparttable}
  \vspace{-0.25cm}
\end{table}

\subsubsection{Inpainter Scalability}
\label{sec:inpaint-impl-scale}

Though the Free Form model has better accuracy at higher resolutions than other tested methods, it is fairly memory and compute intensive when trained on high-resolution images (600x600) and used for inference on very high-resolution images (1200x1200). In this section we describe methods of decreasing the memory and computational load. 

\begin{figure}
  \begin{subfigure}{\linewidth}
        \includegraphics[width=0.45\linewidth]{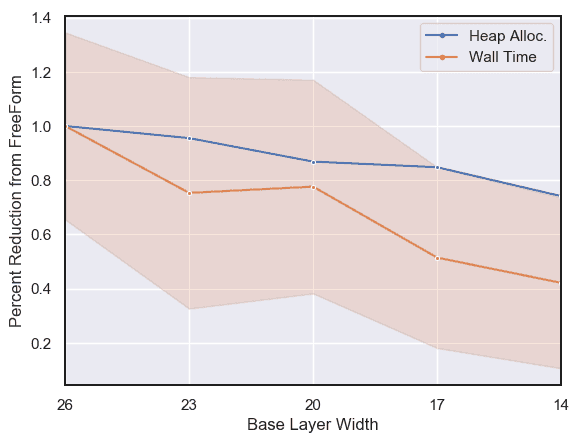}
        \includegraphics[width=0.45\linewidth]{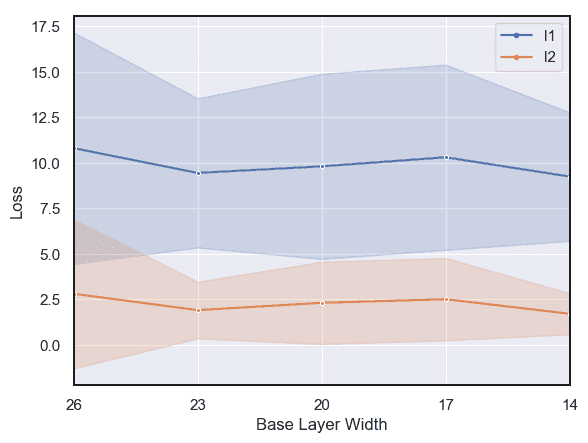}
        \caption{Base Layer Width}
    \end{subfigure}

    \begin{subfigure}{\linewidth}
        \includegraphics[width=0.45\linewidth]{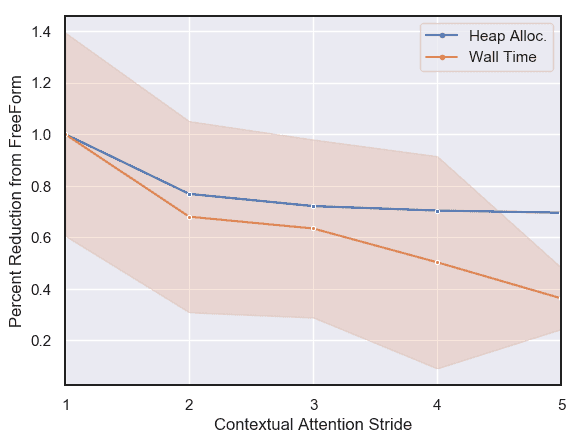}
      \includegraphics[width=0.45\linewidth]{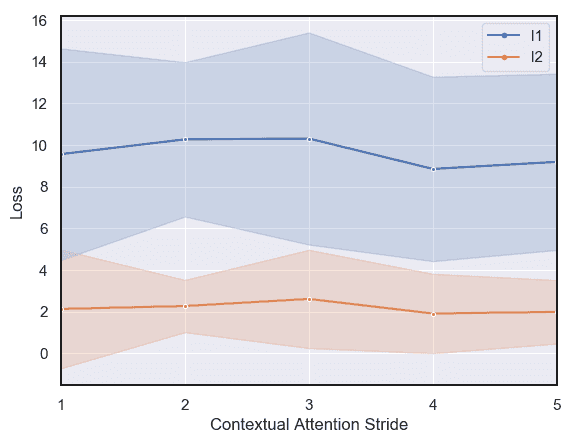}
        \caption{Contextual Attention Stride}
    \end{subfigure}
	\caption{Measurement of scalability and loss metrics over changing hyperparameters.}
	\Description{Measurement of scalability and loss metrics over changing model width.}
	\label{fig:hyperparams}
\end{figure}

Given an image size, the two hyperparameters that have the biggest impact on computational cost are the base layer width (all layers in the model are a multiple of this hyperparameter) and the stride of the contextual attention layer. Both of these hyperparameters relate to model capacity; reducing capacity likely impacts the quality of the inpainting model. To examine this relationship, we separately vary these two hyperparameters and measure the quantitative loss scores and run-time metrics in Figure \ref{fig:hyperparams}. As expected, decreasing the base layer width results in less heap allocation and less wall time usage, as there are less parameters in the model. Increasing the stride of the contextual attention layer has a similar effect, although we note that the decrease in allocated memory levels off. We expected $l1$ and $l2$ loss to increase as model capacity decreases. Instead, we observe a slight trend in the \textit{opposite} direction. We note that there are extremely high standard deviations, making it difficult to draw meaningful conclusions from the loss metrics. In accordance with our original hypothesis, we observe significant visual degradation in qualitative tasks when using hyperparameter settings that result in decreased model capacity, despite similar loss values. We believe this further suggests that $l1$ and $l2$ loss have a low correlation to inpainting quality. Based on our overall observations and our run-time measurements, we select a base layer width of 20 and a contextual attention stride of two\footnote{Compared to baseline values of 26 and one respectively.}. 

Inpainting images larger than 1200x1200px is challenging even with decreased model size. To solve this, we slice the image around each separated mask component and inpaint each slice separately. We then stitch the results back together. We call this approach `Low Memory' Inpainting. We analyze the percent reduction in memory and in run-time in Table \ref{tab:inpainter-scale}. Here, `Nostalgin' refers to a Free Form inpainting model with the hyperparameter changes discussed above. Note that we do not report the standard deviation for heap allocation; to calculate heap allocation, we could only easily measure the final heap across our evaluation set. We divide that value by the number of evaluation images. With our changes to model width and contextual attention stride, our model performs more than 50\% faster and uses almost 30\% less heap allocation than the one proposed by \cite{yu2018free}. When combining the Low Memory Inpainter we achieve more than 90\% wall time speed up, using nearly 80\% less heap allocation. This speedup comes with only slight increases in $l1$ and $l2$ loss, and (qualitatively) a few additional visual artifacts.

\begin{table}
  \begin{threeparttable}
    \caption{\% Reduction in Inpainting Scalability Metrics}
    \label{tab:inpainter-scale}
    \begin{tabular}{lccc}
        \toprule
        & Wall Time \hfill & Heap Alloc. \\
        \midrule
        Nostalgin (Full Image) & 54.7 $\pm$ 4.6 & 27.8  \\
        Nostalgin (Low Memory) & 90.7 $\pm$ 3.1 & 79.6 \\
        \bottomrule
    \end{tabular}
    \begin{center}
    \textit{All percentages compared to the FreeForm method \cite{yu2018free}.}    
\end{center}
  \end{threeparttable}
  \vspace{-0.5cm}
\end{table}

\subsection{Modeling}
\label{sec:model-impl}

\begin{figure}
    \begin{subfigure}{0.12\textwidth}
        \includegraphics[width=\textwidth]{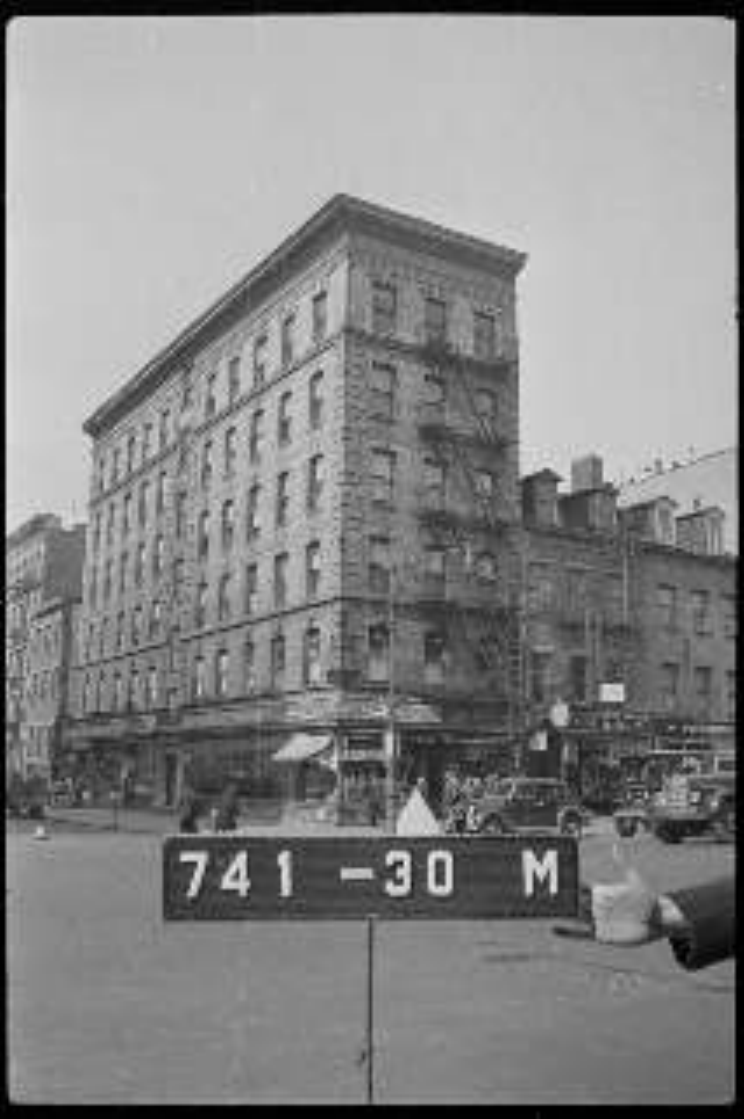}
        \caption{Input.}
    \end{subfigure}
    \begin{subfigure}{0.256\textwidth}
        \includegraphics[width=0.46875\textwidth]{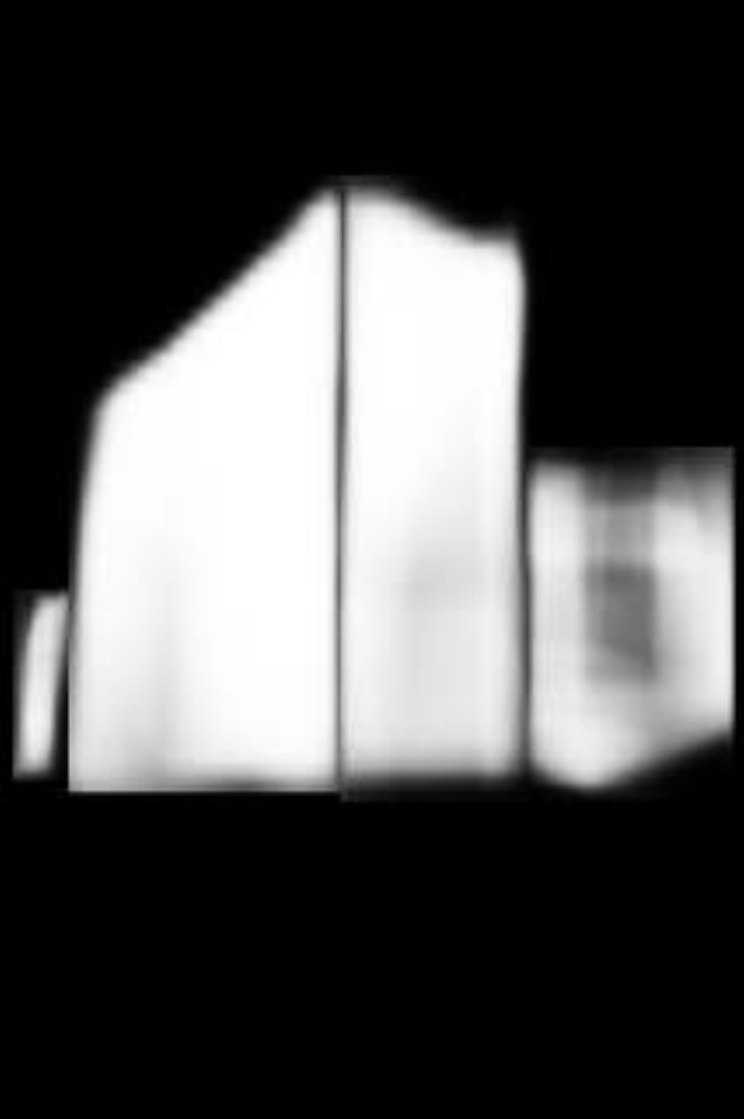}
        \includegraphics[width=0.46875\textwidth]{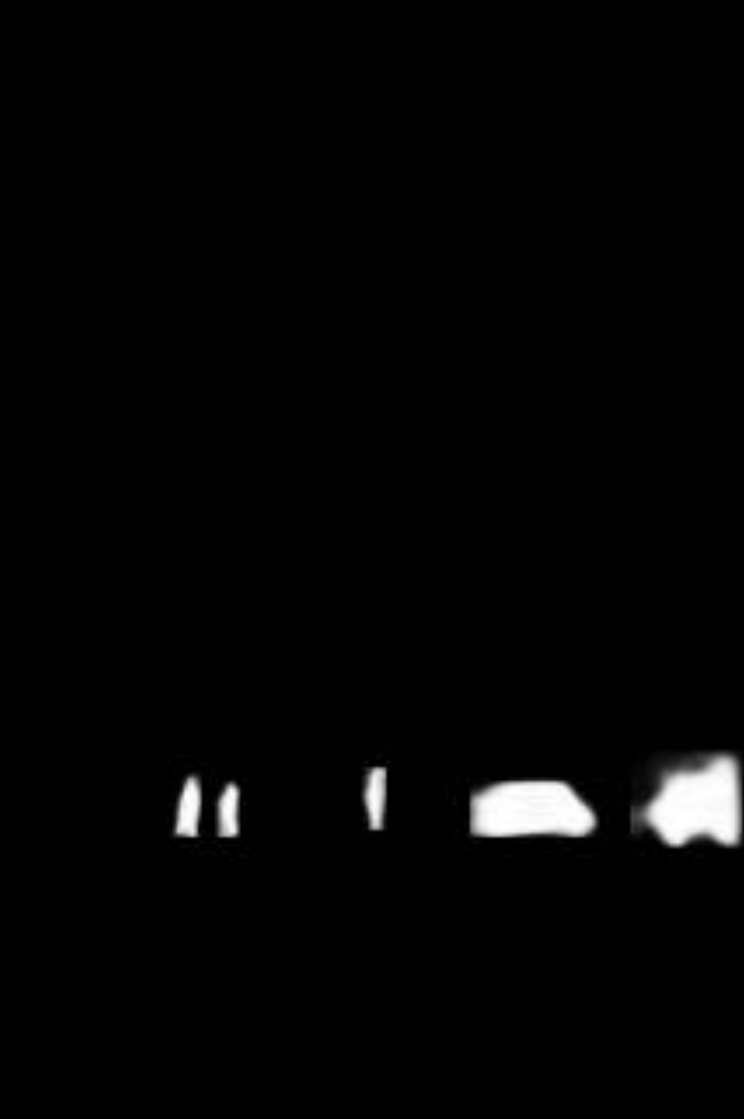}
        \caption{Segment Facade, Occlusions.}
    \end{subfigure}
	
    \begin{subfigure}{0.256\textwidth}
        \includegraphics[width=0.46875\textwidth]{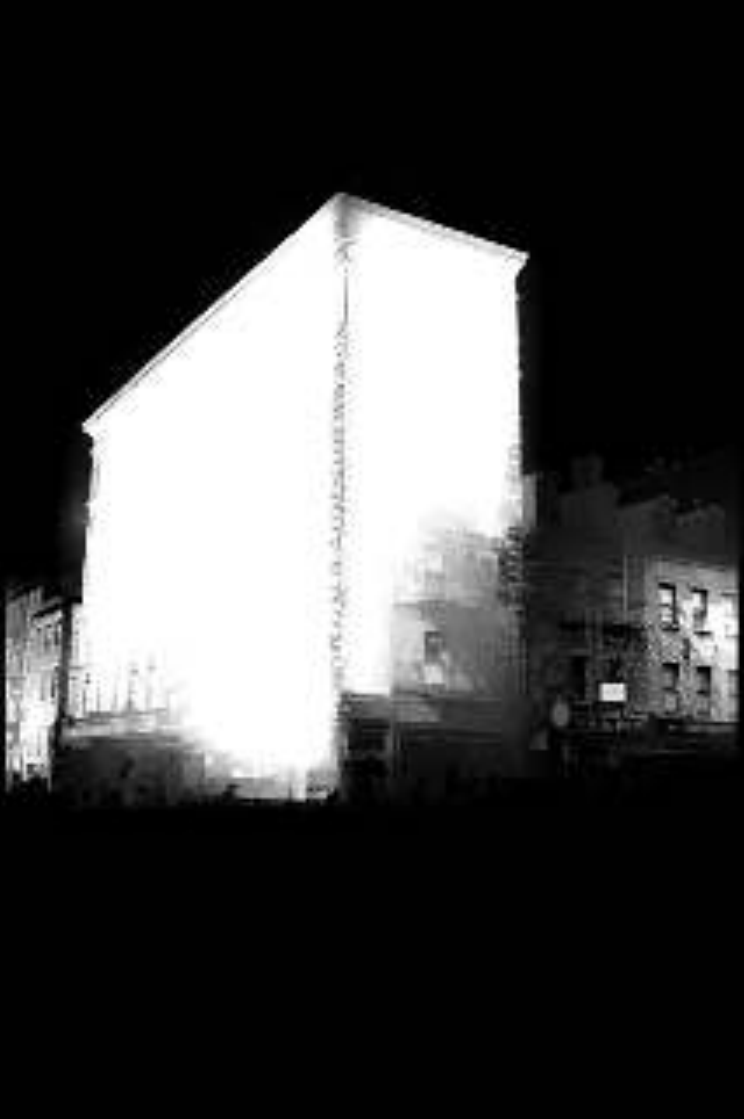}
        \includegraphics[width=0.46875\textwidth]{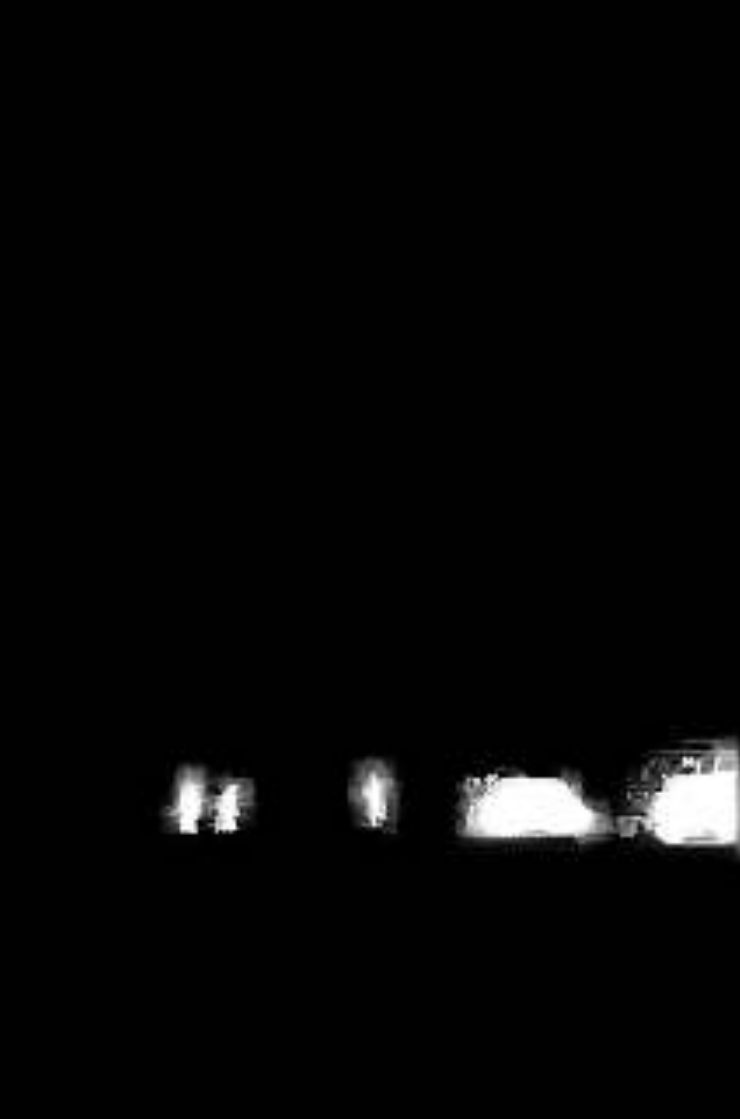}
    \caption{Mat Facade, Occlusions.}
    \end{subfigure}
    
    \begin{subfigure}{0.256\textwidth}
        \includegraphics[width=0.46875\textwidth]{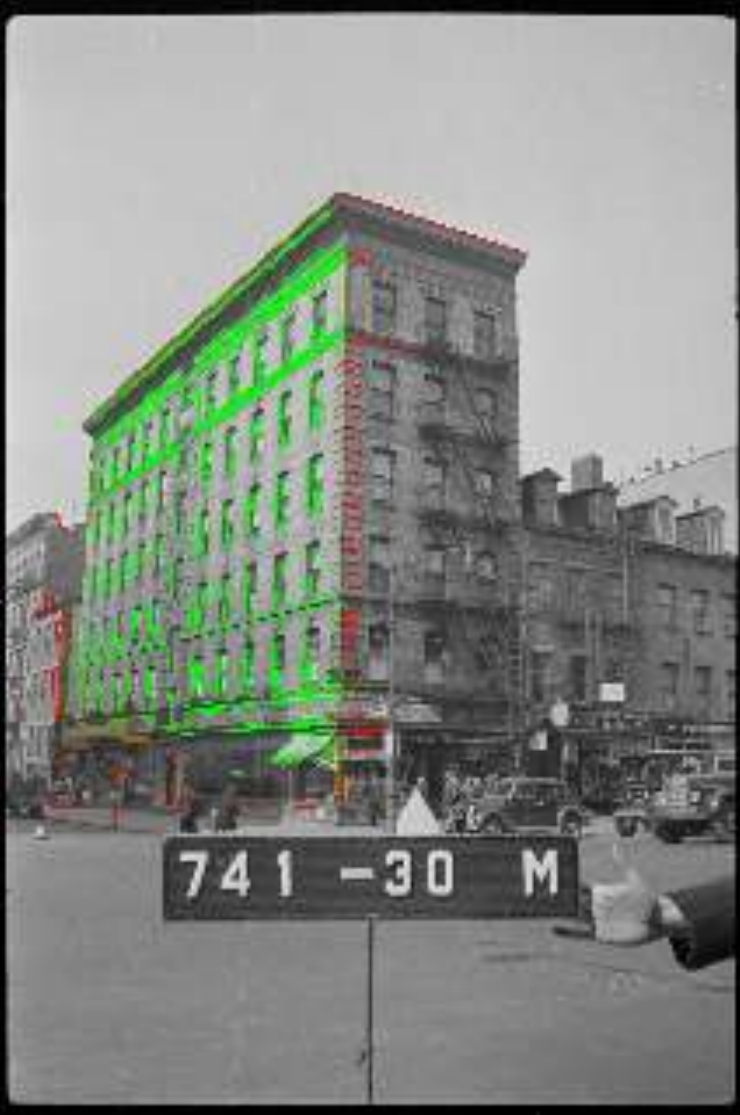}
        \includegraphics[width=0.46875\textwidth]{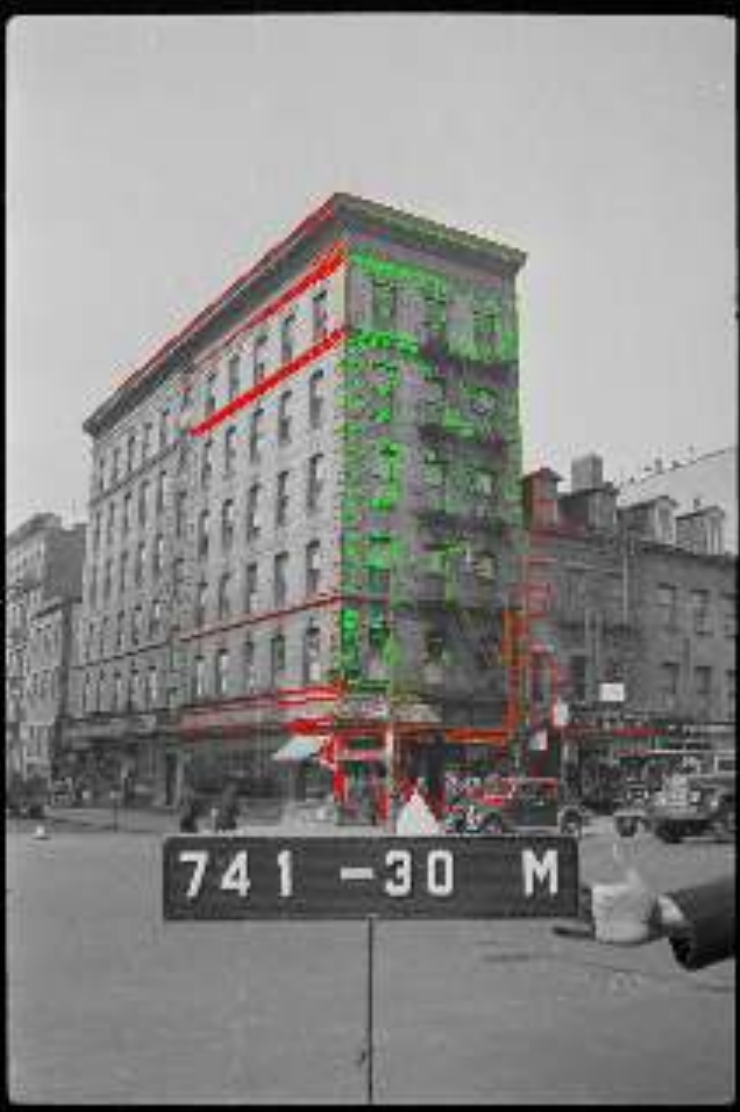}
    \caption{Extract Lines.}
    \end{subfigure}
    
    \begin{subfigure}{0.36\textwidth}
        \includegraphics[width=0.77\textwidth]{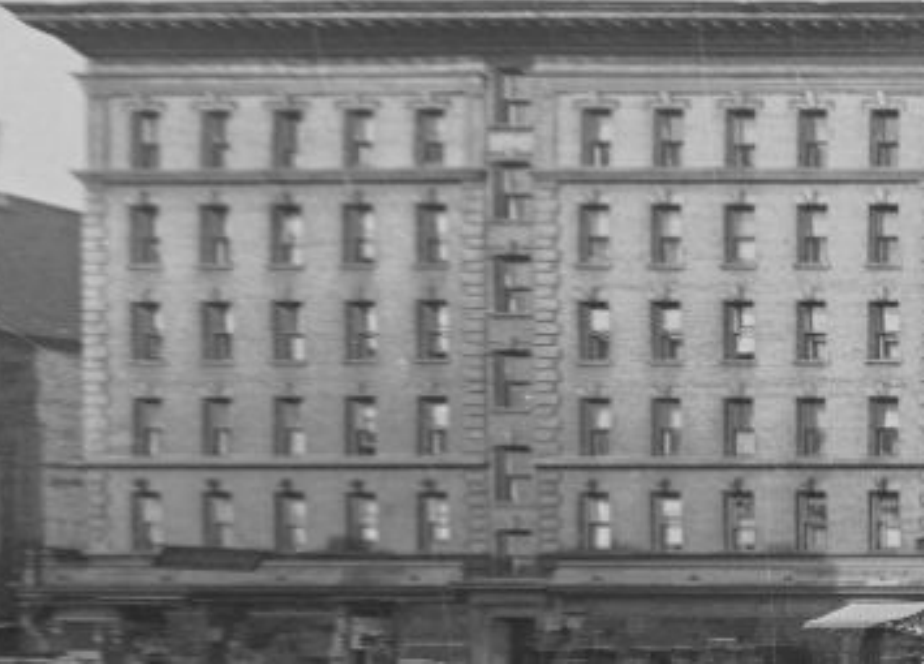}
        \includegraphics[width=0.19\textwidth]{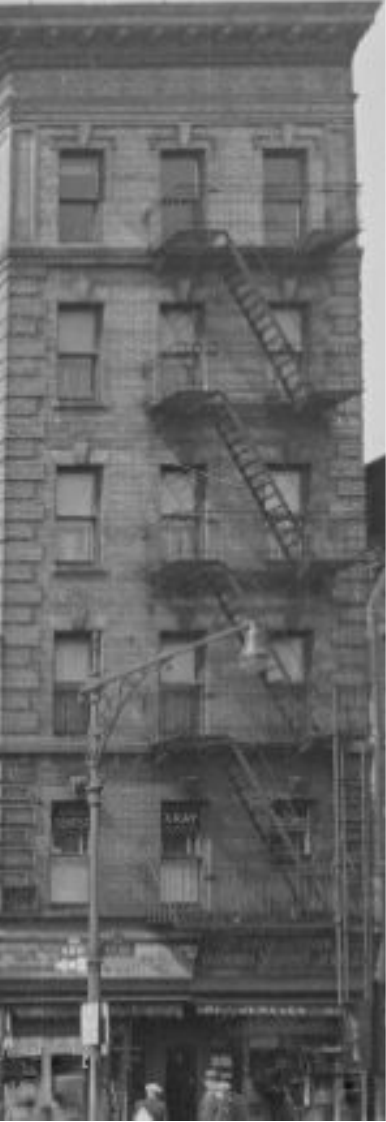}
    \caption{Final Results (Rectify, Inpaint).}
    \end{subfigure}
    
	\caption{End to end processing pipeline depicting 2D facade extraction, rectification, and inpainting. Input image courtesy of New York City Municipal Archives.}
	\Description{End to end processing pipeline depicting 2D facade extraction, rectification, and inpainting. Input image courtesy of New York City Municipal Archives.}
	\label{fig:complete}
\end{figure}

We examine the 3D modeling pipeline end to end by utilizing a set of facade image data to reconstruct two blocks of Manhattan as it looked in the 1940s. The image data for these two blocks are taken from a tax record collection maintained by the New York Municipal Archives. Figure \ref{fig:complete} depicts an input image as it goes through the 2D processing components described above, and demonstrates how clean facades can be extracted. Specifically, we are able to extract two rectified and inpainted facades from a single black and white image of a corner building. 

We are able to run this pipeline at scale for many images in a distributed fashion. We demonstrate this in Figure \ref{fig:3D}, which depicts several angles of our generated city blocks (additional images in the Appendix). We note that the generated environment is fully walkable; the images presented in the figure are screenshots of a larger simulation instead of one-off renderings. Thus, we are able to easily generate viewing angles that are not present in the initial images, showing the power of our approach. We also compare our reconstruction to modern day images taken from Google Streetview. We highlight that several buildings have changed significantly or have completely been removed; as a result, our 3D reconstruction is capable of capturing an experience that no longer exists.

\begin{figure}
    \centering
	\begin{minipage}{0.8\linewidth}
    \textbf{\fixlen{1940s Reconstruction Today}}
        \includegraphics[width=\textwidth]{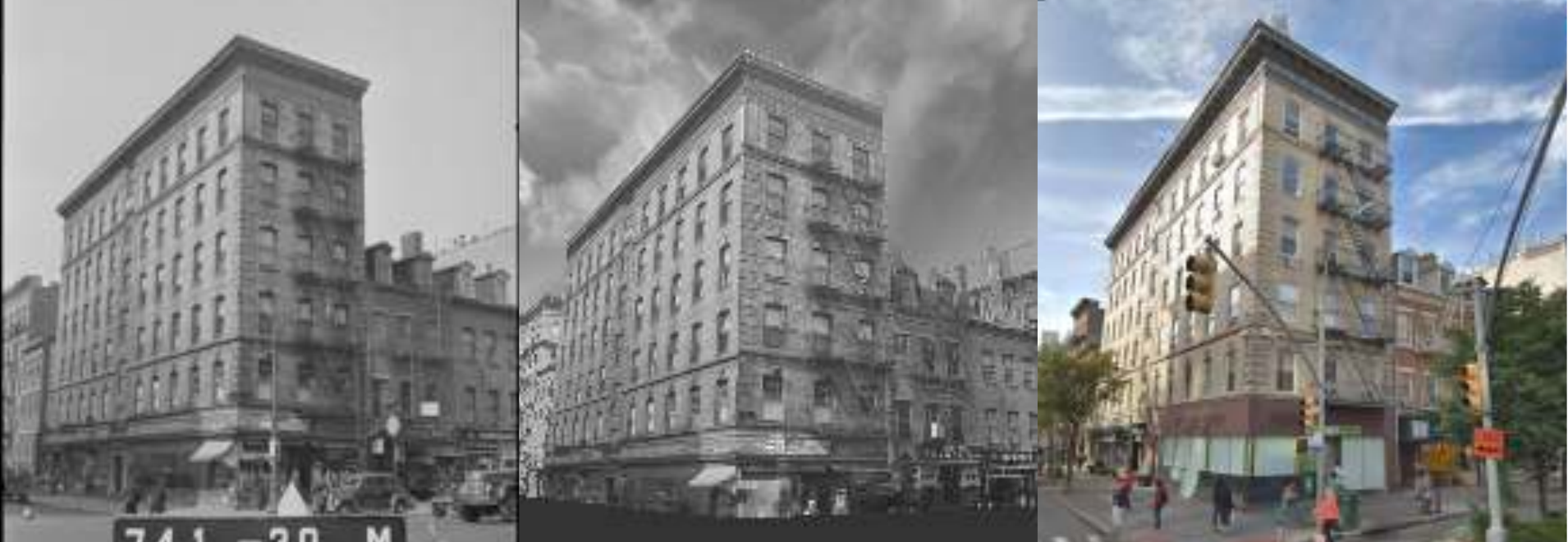}
        \subcaption{7th Ave, 17th St, NW Corner.}
        \includegraphics[width=\textwidth]{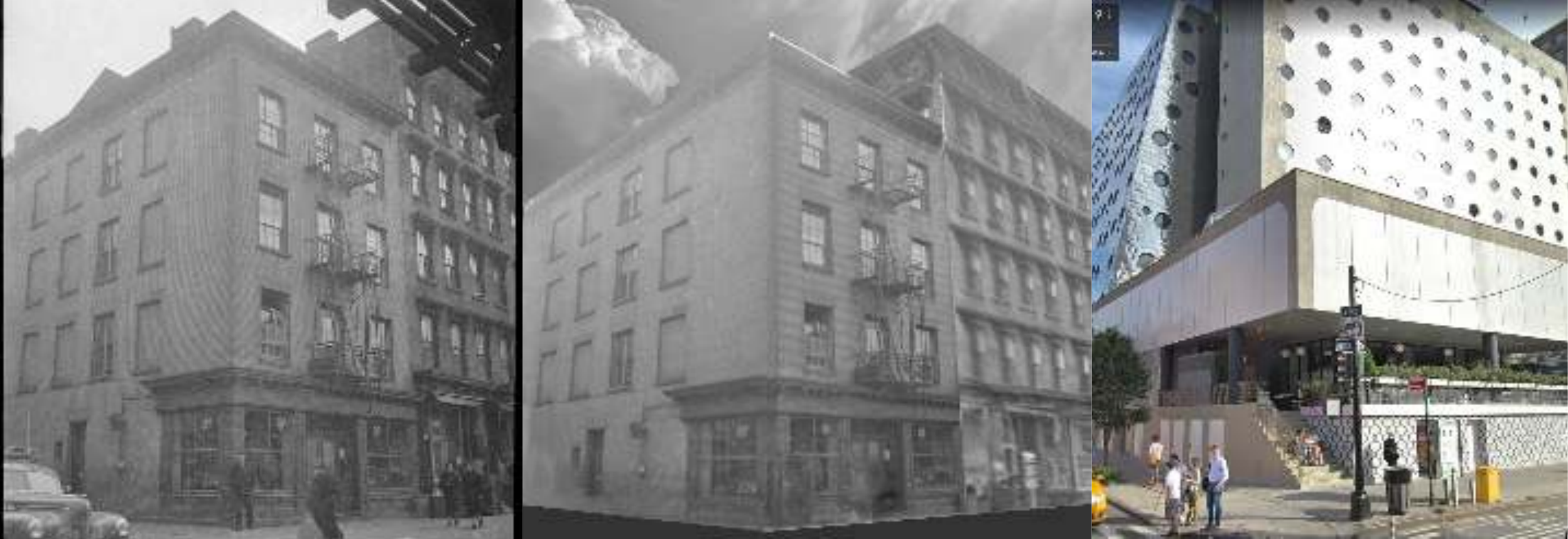}
        \subcaption{9th Ave, 17th St, SE Corner.}
        \includegraphics[width=\textwidth]{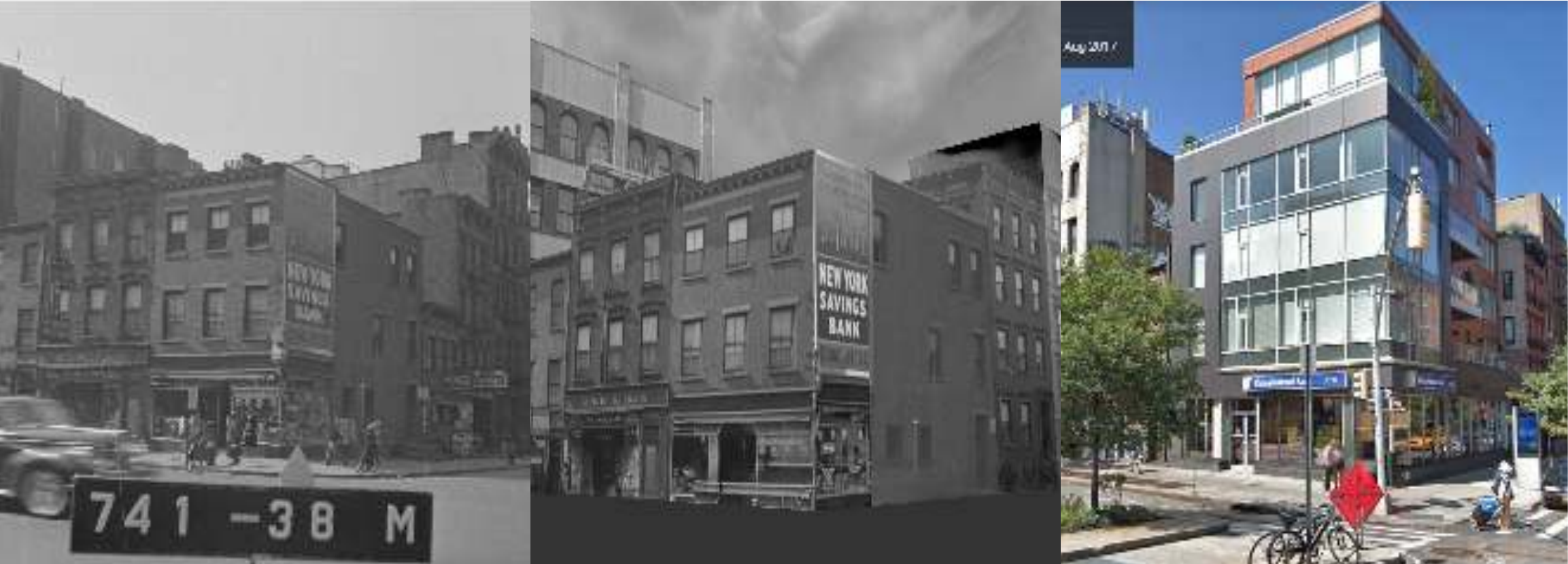}
        \subcaption{9th Ave, 18th St, SW Corner.}
    \end{minipage}
	\caption{Qualitative analysis of inpainting methods. From left to right, we show the original image data (courtesy of the New York City Municipal Archives), our 3D reconstruction, and the modern day (taken from Google Streetview). All images are from New York, NY.}
	\Description{Qualitative analysis of inpainting methods.}
	\vspace{-0.5cm}
	\label{fig:3D}
\end{figure}

\section{Conclusion and Future Work}

Automatic city reconstruction from historical images is a difficult task because historical images provide few guarantees about image quality or content and often do not have important metadata required to extract 3D geometry. In this work, we propose and motivate Nostalgin, a scalable 3D city generator that is specifically built for processing high-resolution historical image data. We describe a four part pipeline composed of image parsing, rectification, inpainting, and modeling. For each component, we examine several design choices and present quantitative and qualitative results. We show that each subcomponent is built to uniquely handle the inherent difficulties that arise when dealing with historical image data, such as sparsity of images and lack of metadata. We demonstrate the end-to-end pipeline by reconstructing two Manhattan city blocks from the 1940s.

We aim to leverage the power of Nostalgin to create an open source platform where users can contribute their own photos and generate immersive historical experiences that will allow them to connect to prior eras of history. Additional data collected from such a platform would help us further generalize Nostalgin, helping us move towards full 3D reconstruction of all types of buildings. We also are beginning to examine how we can extract geolocation information from historical plot data, allowing us to move away from any geotagging requirements.

We believe Nostalgin enables users to experience historical settings in a way that was previously impossible. We are excited for future developments in the historical 3D city modeling space. 

%
\begin{acks}

We would like to acknowledge Noah Snavely; without his guidance, this work may not have been done. We would also like to acknowledge Bryan Perozzi, Vahab Mirrokni, Feng Han, and Ameesh Makadia. Finally, we would like to thank the New York City Municipal Archives for their support.

\end{acks}

%
\bibliographystyle{ACM-Reference-Format}
\bibliography{sample-base}


\begin{thebibliography}{28}


\ifx \showCODEN    \undefined \def \showCODEN     #1{\unskip}     \fi
\ifx \showDOI      \undefined \def \showDOI       #1{#1}\fi
\ifx \showISBNx    \undefined \def \showISBNx     #1{\unskip}     \fi
\ifx \showISBNxiii \undefined \def \showISBNxiii  #1{\unskip}     \fi
\ifx \showISSN     \undefined \def \showISSN      #1{\unskip}     \fi
\ifx \showLCCN     \undefined \def \showLCCN      #1{\unskip}     \fi
\ifx \shownote     \undefined \def \shownote      #1{#1}          \fi
\ifx \showarticletitle \undefined \def \showarticletitle #1{#1}   \fi
\ifx \showURL      \undefined \def \showURL       {\relax}        \fi
\providecommand\bibfield[2]{#2}
\providecommand\bibinfo[2]{#2}
\providecommand\natexlab[1]{#1}
\providecommand\showeprint[2][]{arXiv:#2}

\bibitem[\protect\citeauthoryear{Agarwal, Snavely, Seitz, and Szeliski}{Agarwal
  et~al\mbox{.}}{2010}]%
        {agarwal2010bundle}
\bibfield{author}{\bibinfo{person}{Sameer Agarwal}, \bibinfo{person}{Noah
  Snavely}, \bibinfo{person}{Steven~M Seitz}, {and} \bibinfo{person}{Richard
  Szeliski}.} \bibinfo{year}{2010}\natexlab{}.
\newblock \showarticletitle{Bundle adjustment in the large}. In
  \bibinfo{booktitle}{\emph{European conference on computer vision}}. Springer,
  \bibinfo{pages}{29--42}.
\newblock


\bibitem[\protect\citeauthoryear{Agarwal, Snavely, Simon, Seitz, and
  Szeliski}{Agarwal et~al\mbox{.}}{2009}]%
        {agarwal2009building}
\bibfield{author}{\bibinfo{person}{Sameer Agarwal}, \bibinfo{person}{Noah
  Snavely}, \bibinfo{person}{Ian Simon}, \bibinfo{person}{Steven~M Seitz},
  {and} \bibinfo{person}{Richard Szeliski}.} \bibinfo{year}{2009}\natexlab{}.
\newblock \showarticletitle{Building rome in a day}. In
  \bibinfo{booktitle}{\emph{Computer Vision, 2009 IEEE 12th International
  Conference on}}. IEEE, \bibinfo{pages}{72--79}.
\newblock


\bibitem[\protect\citeauthoryear{Barnes, Shechtman, Finkelstein, and
  Goldman}{Barnes et~al\mbox{.}}{2009}]%
        {barnes2009patchmatch}
\bibfield{author}{\bibinfo{person}{Connelly Barnes}, \bibinfo{person}{Eli
  Shechtman}, \bibinfo{person}{Adam Finkelstein}, {and} \bibinfo{person}{Dan~B
  Goldman}.} \bibinfo{year}{2009}\natexlab{}.
\newblock \showarticletitle{PatchMatch: A randomized correspondence algorithm
  for structural image editing}.
\newblock \bibinfo{journal}{\emph{ACM Transactions on Graphics (ToG)}}
  \bibinfo{volume}{28}, \bibinfo{number}{3} (\bibinfo{year}{2009}),
  \bibinfo{pages}{24}.
\newblock


\bibitem[\protect\citeauthoryear{Bertalmio, Bertozzi, and Sapiro}{Bertalmio
  et~al\mbox{.}}{2001}]%
        {bertalmio2001navier}
\bibfield{author}{\bibinfo{person}{Marcelo Bertalmio},
  \bibinfo{person}{Andrea~L Bertozzi}, {and} \bibinfo{person}{Guillermo
  Sapiro}.} \bibinfo{year}{2001}\natexlab{}.
\newblock \showarticletitle{Navier-stokes, fluid dynamics, and image and video
  inpainting}. In \bibinfo{booktitle}{\emph{Computer Vision and Pattern
  Recognition, 2001. CVPR 2001. Proceedings of the 2001 IEEE Computer Society
  Conference on}}, Vol.~\bibinfo{volume}{1}. IEEE, \bibinfo{pages}{I--I}.
\newblock


\bibitem[\protect\citeauthoryear{Chang, Funkhouser, Guibas, Hanrahan, Huang,
  Li, Savarese, Savva, Song, Su, et~al\mbox{.}}{Chang et~al\mbox{.}}{2015}]%
        {chang2015shapenet}
\bibfield{author}{\bibinfo{person}{Angel~X Chang}, \bibinfo{person}{Thomas
  Funkhouser}, \bibinfo{person}{Leonidas Guibas}, \bibinfo{person}{Pat
  Hanrahan}, \bibinfo{person}{Qixing Huang}, \bibinfo{person}{Zimo Li},
  \bibinfo{person}{Silvio Savarese}, \bibinfo{person}{Manolis Savva},
  \bibinfo{person}{Shuran Song}, \bibinfo{person}{Hao Su}, {et~al\mbox{.}}}
  \bibinfo{year}{2015}\natexlab{}.
\newblock \showarticletitle{Shapenet: An information-rich 3d model repository}.
\newblock \bibinfo{journal}{\emph{arXiv preprint arXiv:1512.03012}}
  (\bibinfo{year}{2015}).
\newblock


\bibitem[\protect\citeauthoryear{Fischler and Bolles}{Fischler and
  Bolles}{1981}]%
        {ransac}
\bibfield{author}{\bibinfo{person}{Martin~A. Fischler} {and}
  \bibinfo{person}{Robert~C. Bolles}.} \bibinfo{year}{1981}\natexlab{}.
\newblock \showarticletitle{Random Sample Consensus: A Paradigm for Model
  Fitting with Applications to Image Analysis and Automated Cartography}.
\newblock \bibinfo{journal}{\emph{Commun. ACM}} \bibinfo{volume}{24},
  \bibinfo{number}{6} (\bibinfo{date}{June} \bibinfo{year}{1981}),
  \bibinfo{pages}{381--395}.
\newblock
\showISSN{0001-0782}
\urldef\tempurl%
\url{https://doi.org/10.1145/358669.358692}
\showDOI{\tempurl}


\bibitem[\protect\citeauthoryear{Hara, Vemulapalli, and Chellappa}{Hara
  et~al\mbox{.}}{2017}]%
        {hara2017designing}
\bibfield{author}{\bibinfo{person}{Kota Hara}, \bibinfo{person}{Raviteja
  Vemulapalli}, {and} \bibinfo{person}{Rama Chellappa}.}
  \bibinfo{year}{2017}\natexlab{}.
\newblock \showarticletitle{Designing deep convolutional neural networks for
  continuous object orientation estimation}.
\newblock \bibinfo{journal}{\emph{arXiv preprint arXiv:1702.01499}}
  (\bibinfo{year}{2017}).
\newblock


\bibitem[\protect\citeauthoryear{He, Gkioxari, Dollár, and Girshick}{He
  et~al\mbox{.}}{2017}]%
        {maskrcnn}
\bibfield{author}{\bibinfo{person}{Kaiming He}, \bibinfo{person}{Georgia
  Gkioxari}, \bibinfo{person}{Piotr Dollár}, {and} \bibinfo{person}{Ross
  Girshick}.} \bibinfo{year}{2017}\natexlab{}.
\newblock \showarticletitle{Mask R-CNN}.
\newblock \bibinfo{journal}{\emph{arXiv:1703.06870}} (\bibinfo{year}{2017}).
\newblock


\bibitem[\protect\citeauthoryear{Horry, Anjyo, and Arai}{Horry
  et~al\mbox{.}}{1997}]%
        {horry1997tour}
\bibfield{author}{\bibinfo{person}{Youichi Horry}, \bibinfo{person}{Ken-Ichi
  Anjyo}, {and} \bibinfo{person}{Kiyoshi Arai}.}
  \bibinfo{year}{1997}\natexlab{}.
\newblock \showarticletitle{Tour into the picture: using a spidery mesh
  interface to make animation from a single image}. In
  \bibinfo{booktitle}{\emph{Proceedings of the 24th annual conference on
  Computer graphics and interactive techniques}}. ACM Press/Addison-Wesley
  Publishing Co., \bibinfo{pages}{225--232}.
\newblock


\bibitem[\protect\citeauthoryear{Irschara, Zach, and Bischof}{Irschara
  et~al\mbox{.}}{2007}]%
        {irschara2007towards}
\bibfield{author}{\bibinfo{person}{Arnold Irschara},
  \bibinfo{person}{Christopher Zach}, {and} \bibinfo{person}{Horst Bischof}.}
  \bibinfo{year}{2007}\natexlab{}.
\newblock \showarticletitle{Towards wiki-based dense city modeling}. In
  \bibinfo{booktitle}{\emph{Computer Vision, 2007. ICCV 2007. IEEE 11th
  International Conference on}}. IEEE, \bibinfo{pages}{1--8}.
\newblock


\bibitem[\protect\citeauthoryear{Jiang, Tan, and Cheong}{Jiang
  et~al\mbox{.}}{2009}]%
        {jiang2009symmetric}
\bibfield{author}{\bibinfo{person}{Nianjuan Jiang}, \bibinfo{person}{Ping Tan},
  {and} \bibinfo{person}{Loong-Fah Cheong}.} \bibinfo{year}{2009}\natexlab{}.
\newblock \showarticletitle{Symmetric architecture modeling with a single
  image}.
\newblock \bibinfo{journal}{\emph{ACM Transactions on Graphics (TOG)}}
  \bibinfo{volume}{28}, \bibinfo{number}{5} (\bibinfo{year}{2009}),
  \bibinfo{pages}{113}.
\newblock


\bibitem[\protect\citeauthoryear{Kelly, Femiani, Wonka, and Mitra}{Kelly
  et~al\mbox{.}}{2017}]%
        {kelly2017bigsur}
\bibfield{author}{\bibinfo{person}{Tom Kelly}, \bibinfo{person}{John Femiani},
  \bibinfo{person}{Peter Wonka}, {and} \bibinfo{person}{Niloy~J Mitra}.}
  \bibinfo{year}{2017}\natexlab{}.
\newblock \showarticletitle{BigSUR: large-scale structured urban
  reconstruction}.
\newblock \bibinfo{journal}{\emph{ACM Transactions on Graphics (TOG)}}
  \bibinfo{volume}{36}, \bibinfo{number}{6} (\bibinfo{year}{2017}),
  \bibinfo{pages}{204}.
\newblock


\bibitem[\protect\citeauthoryear{Korah, Medasani, and Owechko}{Korah
  et~al\mbox{.}}{2011}]%
        {korah2011strip}
\bibfield{author}{\bibinfo{person}{Thommen Korah}, \bibinfo{person}{Swarup
  Medasani}, {and} \bibinfo{person}{Yuri Owechko}.}
  \bibinfo{year}{2011}\natexlab{}.
\newblock \showarticletitle{Strip histogram grid for efficient lidar
  segmentation from urban environments}. In \bibinfo{booktitle}{\emph{Computer
  Vision and Pattern Recognition Workshops (CVPRW), 2011 IEEE Computer Society
  Conference on}}. IEEE, \bibinfo{pages}{74--81}.
\newblock


\bibitem[\protect\citeauthoryear{Ko{\v{s}}eck{\'a} and Zhang}{Ko{\v{s}}eck{\'a}
  and Zhang}{2002}]%
        {kovsecka2002video}
\bibfield{author}{\bibinfo{person}{Jana Ko{\v{s}}eck{\'a}} {and}
  \bibinfo{person}{Wei Zhang}.} \bibinfo{year}{2002}\natexlab{}.
\newblock \showarticletitle{Video compass}. In
  \bibinfo{booktitle}{\emph{European conference on computer vision}}. Springer,
  \bibinfo{pages}{476--490}.
\newblock


\bibitem[\protect\citeauthoryear{Ko{\v{s}}eck{\'a} and Zhang}{Ko{\v{s}}eck{\'a}
  and Zhang}{2005}]%
        {kovsecka2005extraction}
\bibfield{author}{\bibinfo{person}{Jana Ko{\v{s}}eck{\'a}} {and}
  \bibinfo{person}{Wei Zhang}.} \bibinfo{year}{2005}\natexlab{}.
\newblock \showarticletitle{Extraction, matching, and pose recovery based on
  dominant rectangular structures}.
\newblock \bibinfo{journal}{\emph{Computer Vision and Image Understanding}}
  \bibinfo{volume}{100}, \bibinfo{number}{3} (\bibinfo{year}{2005}),
  \bibinfo{pages}{274--293}.
\newblock


\bibitem[\protect\citeauthoryear{Levin, Lischinski, and Weiss}{Levin
  et~al\mbox{.}}{2008}]%
        {laplacianmatting}
\bibfield{author}{\bibinfo{person}{Anat Levin}, \bibinfo{person}{Dani
  Lischinski}, {and} \bibinfo{person}{Yair Weiss}.}
  \bibinfo{year}{2008}\natexlab{}.
\newblock \showarticletitle{A Closed Form Solution to Natural Image Matting}.
\newblock \bibinfo{journal}{\emph{IEEE Transactions on Pattern Analysis and
  Machine Intelligence}} \bibinfo{volume}{30}, \bibinfo{number}{2}
  (\bibinfo{year}{2008}), \bibinfo{pages}{228--242}.
\newblock


\bibitem[\protect\citeauthoryear{Liu, Zhang, Zhu, and Hoi}{Liu
  et~al\mbox{.}}{2017}]%
        {liu2017deepfacade}
\bibfield{author}{\bibinfo{person}{Hantang Liu}, \bibinfo{person}{Jialiang
  Zhang}, \bibinfo{person}{Jianke Zhu}, {and} \bibinfo{person}{Steven~CH Hoi}.}
  \bibinfo{year}{2017}\natexlab{}.
\newblock \showarticletitle{Deepfacade: A deep learning approach to facade
  parsing}.
\newblock  (\bibinfo{year}{2017}).
\newblock


\bibitem[\protect\citeauthoryear{Musialski, Wonka, Aliaga, Wimmer, Van~Gool,
  and Purgathofer}{Musialski et~al\mbox{.}}{2013}]%
        {musialski2013survey}
\bibfield{author}{\bibinfo{person}{Przemyslaw Musialski},
  \bibinfo{person}{Peter Wonka}, \bibinfo{person}{Daniel~G Aliaga},
  \bibinfo{person}{Michael Wimmer}, \bibinfo{person}{Luc Van~Gool}, {and}
  \bibinfo{person}{Werner Purgathofer}.} \bibinfo{year}{2013}\natexlab{}.
\newblock \showarticletitle{A survey of urban reconstruction}. In
  \bibinfo{booktitle}{\emph{Computer graphics forum}},
  Vol.~\bibinfo{volume}{32}. Wiley Online Library, \bibinfo{pages}{146--177}.
\newblock


\bibitem[\protect\citeauthoryear{Nishida, Bousseau, and Aliaga}{Nishida
  et~al\mbox{.}}{2018}]%
        {nishida2018procedural}
\bibfield{author}{\bibinfo{person}{Gen Nishida}, \bibinfo{person}{Adrien
  Bousseau}, {and} \bibinfo{person}{Daniel~G Aliaga}.}
  \bibinfo{year}{2018}\natexlab{}.
\newblock \showarticletitle{Procedural Modeling of a Building from a Single
  Image}. In \bibinfo{booktitle}{\emph{Computer Graphics Forum}},
  Vol.~\bibinfo{volume}{37}. Wiley Online Library, \bibinfo{pages}{415--429}.
\newblock


\bibitem[\protect\citeauthoryear{Oh, Chen, Dorsey, and Durand}{Oh
  et~al\mbox{.}}{2001}]%
        {oh2001image}
\bibfield{author}{\bibinfo{person}{Byong~Mok Oh}, \bibinfo{person}{Max Chen},
  \bibinfo{person}{Julie Dorsey}, {and} \bibinfo{person}{Fr{\'e}do Durand}.}
  \bibinfo{year}{2001}\natexlab{}.
\newblock \showarticletitle{Image-based modeling and photo editing}. In
  \bibinfo{booktitle}{\emph{Proceedings of the 28th annual conference on
  Computer graphics and interactive techniques}}. ACM,
  \bibinfo{pages}{433--442}.
\newblock


\bibitem[\protect\citeauthoryear{Rother}{Rother}{2002}]%
        {rothervps}
\bibfield{author}{\bibinfo{person}{Carsten Rother}.}
  \bibinfo{year}{2002}\natexlab{}.
\newblock \showarticletitle{A new approach to vanishing point detection in
  architectural environments}.
\newblock \bibinfo{journal}{\emph{Image and Vision Computing}}
  \bibinfo{volume}{20}, \bibinfo{number}{9--10} (\bibinfo{year}{2002}),
  \bibinfo{pages}{647--655}.
\newblock


\bibitem[\protect\citeauthoryear{Vanegas, Aliaga, Wonka, M{\"u}ller, Waddell,
  and Watson}{Vanegas et~al\mbox{.}}{2010}]%
        {vanegas2010modelling}
\bibfield{author}{\bibinfo{person}{Carlos~A Vanegas}, \bibinfo{person}{Daniel~G
  Aliaga}, \bibinfo{person}{Peter Wonka}, \bibinfo{person}{Pascal M{\"u}ller},
  \bibinfo{person}{Paul Waddell}, {and} \bibinfo{person}{Benjamin Watson}.}
  \bibinfo{year}{2010}\natexlab{}.
\newblock \showarticletitle{Modelling the appearance and behaviour of urban
  spaces}. In \bibinfo{booktitle}{\emph{Computer Graphics Forum}},
  Vol.~\bibinfo{volume}{29}. Wiley Online Library, \bibinfo{pages}{25--42}.
\newblock


\bibitem[\protect\citeauthoryear{von Gioi, Jakubowicz, Morel, and Randall}{von
  Gioi et~al\mbox{.}}{2010}]%
        {lsd}
\bibfield{author}{\bibinfo{person}{Rafael~Grompone von Gioi},
  \bibinfo{person}{Jeremie Jakubowicz}, \bibinfo{person}{Jean-Michel Morel},
  {and} \bibinfo{person}{Gregory Randall}.} \bibinfo{year}{2010}\natexlab{}.
\newblock \showarticletitle{LSD: A Fast Line Segment Detector with a False
  Detection Control}.
\newblock \bibinfo{journal}{\emph{IEEE Transactions on Pattern Analysis and
  Machine Intelligence}} \bibinfo{volume}{32}, \bibinfo{number}{4}
  (\bibinfo{year}{2010}), \bibinfo{pages}{722--732}.
\newblock


\bibitem[\protect\citeauthoryear{Yeh, Chen, Yian~Lim, Schwing,
  Hasegawa-Johnson, and Do}{Yeh et~al\mbox{.}}{2017}]%
        {yeh2017semantic}
\bibfield{author}{\bibinfo{person}{Raymond~A Yeh}, \bibinfo{person}{Chen Chen},
  \bibinfo{person}{Teck Yian~Lim}, \bibinfo{person}{Alexander~G Schwing},
  \bibinfo{person}{Mark Hasegawa-Johnson}, {and} \bibinfo{person}{Minh~N Do}.}
  \bibinfo{year}{2017}\natexlab{}.
\newblock \showarticletitle{Semantic image inpainting with deep generative
  models}. In \bibinfo{booktitle}{\emph{Proceedings of the IEEE Conference on
  Computer Vision and Pattern Recognition}}. \bibinfo{pages}{5485--5493}.
\newblock


\bibitem[\protect\citeauthoryear{Yin, Wonka, and Razdan}{Yin
  et~al\mbox{.}}{2009}]%
        {yin2009generating}
\bibfield{author}{\bibinfo{person}{Xuetao Yin}, \bibinfo{person}{Peter Wonka},
  {and} \bibinfo{person}{Anshuman Razdan}.} \bibinfo{year}{2009}\natexlab{}.
\newblock \showarticletitle{Generating 3d building models from architectural
  drawings: A survey}.
\newblock \bibinfo{journal}{\emph{IEEE computer graphics and applications}}
  \bibinfo{volume}{29}, \bibinfo{number}{1} (\bibinfo{year}{2009}).
\newblock


\bibitem[\protect\citeauthoryear{Yu, Lin, Yang, Shen, Lu, and Huang}{Yu
  et~al\mbox{.}}{2018a}]%
        {yu2018free}
\bibfield{author}{\bibinfo{person}{Jiahui Yu}, \bibinfo{person}{Zhe Lin},
  \bibinfo{person}{Jimei Yang}, \bibinfo{person}{Xiaohui Shen},
  \bibinfo{person}{Xin Lu}, {and} \bibinfo{person}{Thomas~S Huang}.}
  \bibinfo{year}{2018}\natexlab{a}.
\newblock \showarticletitle{Free-Form Image Inpainting with Gated Convolution}.
\newblock \bibinfo{journal}{\emph{arXiv preprint arXiv:1806.03589}}
  (\bibinfo{year}{2018}).
\newblock


\bibitem[\protect\citeauthoryear{Yu, Lin, Yang, Shen, Lu, and Huang}{Yu
  et~al\mbox{.}}{2018b}]%
        {yu2018generative}
\bibfield{author}{\bibinfo{person}{Jiahui Yu}, \bibinfo{person}{Zhe Lin},
  \bibinfo{person}{Jimei Yang}, \bibinfo{person}{Xiaohui Shen},
  \bibinfo{person}{Xin Lu}, {and} \bibinfo{person}{Thomas~S Huang}.}
  \bibinfo{year}{2018}\natexlab{b}.
\newblock \showarticletitle{Generative image inpainting with contextual
  attention}.
\newblock \bibinfo{journal}{\emph{arXiv preprint}} (\bibinfo{year}{2018}).
\newblock


\bibitem[\protect\citeauthoryear{Zhang}{Zhang}{[n. d.]}]%
        {zhangwhiteboard}
\bibfield{author}{\bibinfo{person}{Zhengyou Zhang}.} \bibinfo{year}{[n.
  d.]}\natexlab{}.
\newblock \bibinfo{title}{Single-View Geometry of A Rectangle With Application
  to Whiteboard Image Rectification}.
\newblock
\newblock


\end{thebibliography}

%
\clearpage

\appendix

\section{Reproducibility}
\label{sec:supplement}

Below, we describe the settings and hyperparameters used for each component in Nostalgin as well as for experiments described in this paper. All deep learning models are implemented and trained using Tensorflow version 1.12. All run-time results are measured using C++ implementations (Tensorflow models are exported and weights are reloaded in C++ wrappers). 

\subsection{Settings for 3D Manhattan Visualization/Final Settings for Nostalgin}

\subsubsection{Segmentation and Matting}

We trained separate MaskRCNN models for detecting occlusions and for detecting facades. 

For occlusions, we utilized a FasterRCNN inception resnet v3 model\footnote{See: \url{https://github.com/tensorflow/models/blob/master/research/object_detection/models/faster_rcnn_inception_resnet_v2_feature_extractor.py}} trained on the COCO dataset.  The MaskRCNN first stage is trained with four scales of [0.25, 0.5, 1.0, 2.0] and three aspect ratios [0.5, 1.0, 2.0], with a height and width stride of 16. We set the first stage IoU threshold to 0.7. The second stage is trained with four convolutional layers. The mask height and width is 33 by 33. We train with a momentum optimizer, with momentum set to 0.9. We utilize a manual step learning rate that degrades from 3e-4 to 3e-5 at 900k steps, and from 3e-5 to 3e-6 by 1.2M steps.

For facades, we utilized a FasterRCNN inception v2 model\footnote{See: \url{https://github.com/tensorflow/models/blob/master/research/object_detection/models/faster_rcnn_inception_v2_feature_extractor.py}} that is pretrained on COCO and then fine tuned  on roughly 30.5k images of buildings with random horizontal flip, each with a mask drawn around one or more facades present in the image. The first stage had similar parameters to the Occlusion MaskRCNN. The second stage is trained with two convolutional layers. The mask height and width is 46 by 46. We train with the Adam optimizer. We utilize a manual step learning rate that degrades from 5e-5 to 2e-6 between steps 0 and 120000, and from 2e-6 to 1e-6 between steps 1200000 to 220000. We clip gradients to 10.0. 

For both models, batch size is 1, and gradients are clipped to 10.0.

\subsubsection{Rectification}
\label{sec:rectification-supplement}

For rectification, we use Canny thresholds using $\lambda=0.33$ and vanishing point voting alignment uses $t_a=2deg$. When accumulating vanishing points, line detection uses $k_s=16, t_\alpha=4deg$; however, when voting, $k_s=8, t_\alpha=9deg$ is used. We note that determining whether or not an image is rectified is not an easy task; as a result, these hyperparameters were selected based on qualitative assessment of rectification outcomes.

We use different line detection parameters for vanishing point accumulation and voting in order to further reduce the search space of vanishing points. By using more strict linearity parameters, we ensure that vanishing point candidates are only formed using the best straight-line candidates. Whereas during voting we use smaller, low signal lines in order to maximize signal from the image. In cases where we cannot obtain more than 2 vanishing points candidates during accumulation, we relax our parameters to the same values used by voting. We define the local linearity algorithm in Algorithm \ref{algo:linearity}.

\begin{algorithm}
\caption{Local linearity}
\label{algo:linearity}
\begin{algorithmic}
\Procedure{Linearity}{$C[]$}
\State $oncurve \gets \texttt{False}$
\For{$i \gets k_s-1$ to $1$}
  \State $oncurve \gets \texttt{not}$ $oncurve$ and $R_\alpha(C, C_i) < t_\alpha$
\EndFor
\State $oncurve \gets \texttt{True}$
\For{$i \gets k_s$ to $\Call{length}{C}-k_s$}
  \If{$oncurve$ and $R_\alpha(C, C_i) < t_\alpha$}
    \State $oncurve \gets \texttt{False}$
  \EndIf
  \If{$\texttt{not}$ $oncurve$ and $L_\alpha(C, C_i) >= t_\alpha$}
    \State $oncurve \gets \texttt{True}$
  \EndIf
\EndFor
\State $oncurve \gets \texttt{False}$
\For{$i \gets \Call{length}{C}-k_s$ to $\Call{length}{C}$}
  \State $oncurve \gets oncurve$ or $L_\alpha(C, C_i) >= t_\alpha$
\EndFor
\EndProcedure
\end{algorithmic}
\end{algorithm}

\subsubsection{Inpainting}
\label{app-inpainting}

For the inpainter, we follow the structure of \cite{yu2018free}. We set our base model width to 20 instead of 26 in the generator. We set the stride for the contextual attention layer to 2. We set the kernel size for the first three layers of the generator to 20, 10, and 5 respectively. Finally, we clip all gradients to 1.0. 

In order to train our inpainter for the historical image modeling task, we collect a large internal image dataset of about 10M images, each with at least one facade in the image and each having a minimum side length of at least 1000px. The images in this dataset are converted to black and white. The model is trained on 400px by 600px random crops of these images with a batch size of 8 using an Adam optimizer with a learning rate of 2e-5. 

\subsubsection{Modeling}

Image data for our final visualization is provided by the New York Municipal Archives. Renderings are done using Three.js and WebGL. Rough estimations of relative widths are extracted manually from Google Maps building footprints and from the image data. Location data is extracted using OCR on the sign in each image to determine the lot and image number, which is then cross referenced against geotagged references provided by the New York Municipal Archives. For placing images relative to each other, we refer to Algorithm \ref{algo:location}.

\begin{algorithm}
\caption{Mapping Facades to Locations Within a Block}
\label{algo:location}
\begin{algorithmic}
\Procedure{MapFacadesWithinBlock}{$startImage$}
    \State $x \gets 0$
    \State $y \gets 0$

    \State $pointer \gets startImage$
    \State $cardinal \gets startImage.cardinal$
    \While{$pointer.neighbor \neq startImage$}
        \State $height \gets pointer.height$
        \State $alt \gets pointer.length$
        \State $x' \gets 0$
        \State $y' \gets 0$
        \If{$\Call{SameBuilding}{pointer, pointer.neighbor}$}
            \State $alt \gets pointer.neighbor.length$
            \State $cardinal \gets (cardinal + 1)\mod 4$
        \EndIf
        \If{$cardinal = 0$}
            \State $width \gets pointer.length$
            \State $depth \gets alt$
            \State $x' \gets width$
        \ElsIf{$cardinal = 1$}
            \State $depth \gets pointer.length$
            \State $width \gets alt$
            \State $y' \gets depth$
        \ElsIf{$cardinal = 2$}
            \State $width \gets pointer.length$
            \State $depth \gets alt$
            \State $x' \gets -width$
        \Else
            \State $depth \gets pointer.length$
            \State $width \gets alt$
            \State $y' \gets -depth$
        \EndIf
        
        \State $\Call{CreateCube}{height, width, depth, x, y}$
        \State $x = x + x'$
        \State $y = y + y'$
    \EndWhile
\EndProcedure
\end{algorithmic}
\end{algorithm}

\subsection{Settings for Experimental Results}

\subsubsection{Segmentation and Matting}

Precision and recall results for segmentation and matting are calculated as a moving average across 1024 images of roughly 1200x800 pixel resolution. Each image had at least one facade present. For MaskRCNN, the threshold for selecting a pixel as part of the facade is set to 0.5 (i.e. if the model is more than 50\% confident that the pixel is part of the facade, treat the pixel as true). For MaskRCNN with Matting, the threshold for selecting a pixel is set to 0.1. This is lowered due to the nature of what these thresholds represent: in the case of MaskRCNN it is a probability whereas in the case of matting it is the alpha mixing coefficient $\alpha_i$ as defined by
\begin{equation}I_i=\alpha_iF_i+(1-\alpha_i)B_i \quad \alpha_i \in [0, 1]\end{equation}
where $B_i$ and $F_i$ represent the foreground and background colors respectively and $\alpha_i$ represents the mixing coefficient between the two for pixel $I_i$.

The facades in each image in the evaluation set are labelled manually. We note that the ground truth in each image did not necessarily cover \textit{all} facades in the building; thus, there are cases where the segmenter would pick up on a real facade that is not in the ground truth.

\subsubsection{Rectification}

For our analysis of vanishing point space reduction, we run our rectification subcomponent on 1024 images taken from the 10M dataset. These images were all resized to have a minimum side length of 2048px. All rectification hyperparameters are the same as in \ref{sec:rectification-supplement}.

\subsubsection{Inpainting}

Our inpainting qualitative experiment is done on a historical image provided by the New York Municipal Archives. The image size is roughly 1200x800px. The Navier Stokes method is run using the OpenCV v3.4.2 inpaint method, with an inpaint radius of 3 pixels. PatchMatch is run using a minimum patch size of 50px, a maximum patch size of 73px, and a search area size of 100px. For our qualitative measure, all deep models are trained on the 10M dataset. For consistency with prior work, all quantitative metrics are evaluated on the Places2 dataset. Note that though our models are evaluated on Places2, we train our models on the 10M dataset.

When comparing against other methods in Table \ref{tab:inpainter-quant}, we evaluate the inpainter described in \ref{app-inpainting} on 1024 images taken randomly from Places2. The low-memory inpainter uses a maximum chunk size of 400x600px, and context radius of 100px.

For our layer width and stride experiments, we train each inpainter on 200x300px images from the 10M dataset. The inpainters are trained for 1M steps with a batch size of 16 on the 10M historical image set. The models are \textit{not} trained to convergence. We examined quantitative results by measuring the mean $l1$ and $l2$ loss of inpainted results on a held-out set of 1024 images from the 10M set, using randomly drawn free-form masks (see \cite{yu2018free} for the free-form mask algorithm). All percentages are calculated with respect to the average run time of the baseline Free Form method. 

\clearpage
\section{Additional Results}

\begin{figure*}[b]
    \begin{subfigure}{0.36\linewidth}
    \includegraphics[width=\linewidth]{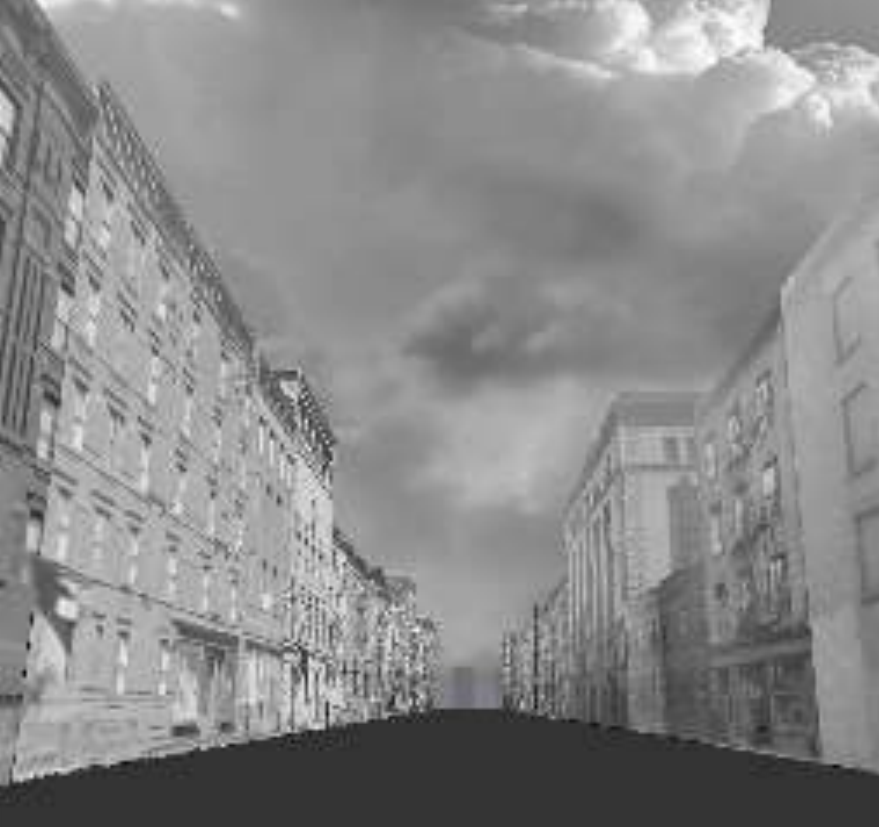}
    \caption{17th Street, facing East.}
    \end{subfigure}
    \hfill
    \begin{subfigure}{0.61\linewidth}
    \includegraphics[width=\linewidth]{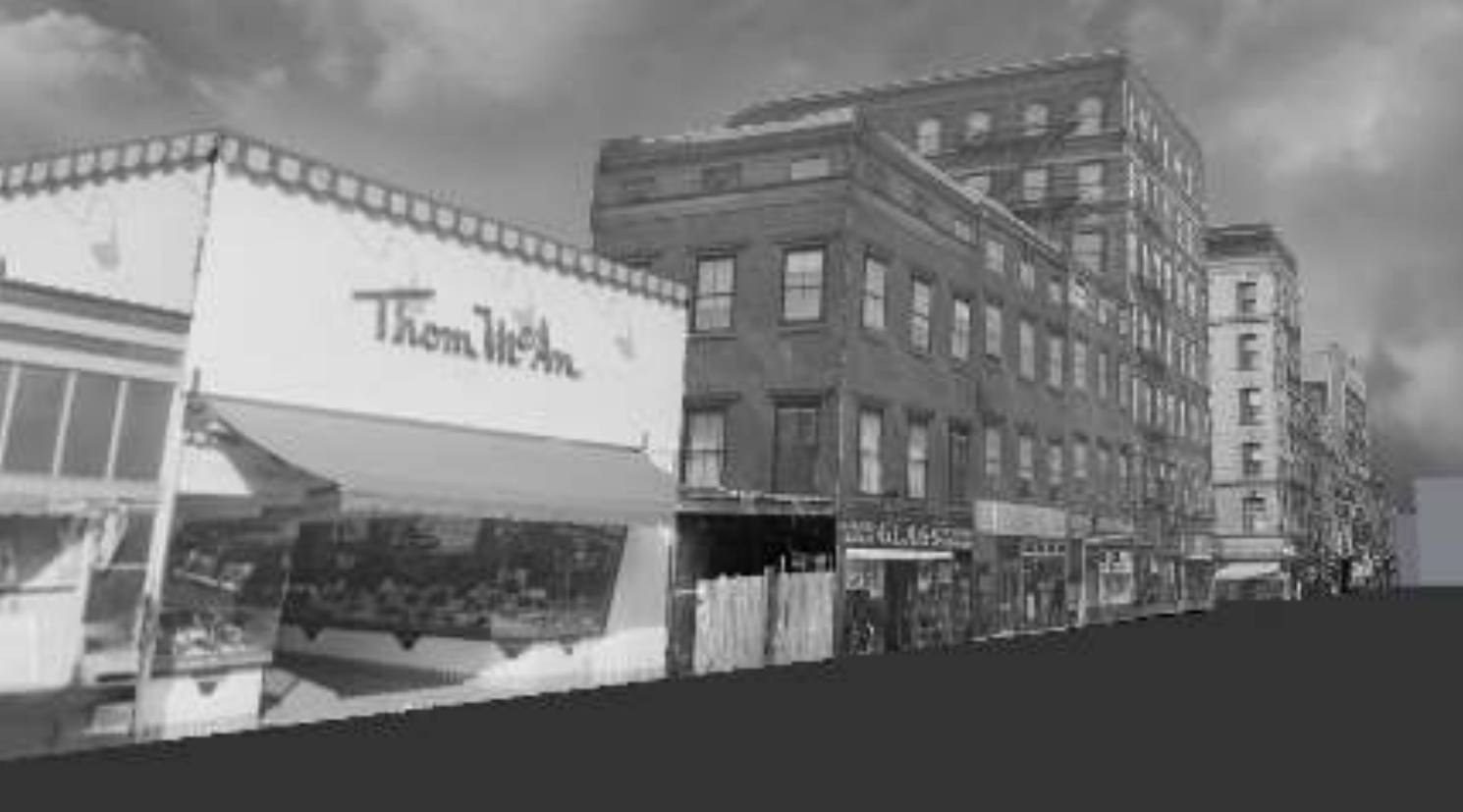}
    \caption{8th Avenue, facing North.}
    \end{subfigure}
    
    \begin{subfigure}{\linewidth}
    \includegraphics[width=\linewidth]{figs/app_9th_16th_ne.pdf}
    \caption{9th Avenue, 16th Street, NE Corner.}
    \end{subfigure}
    
    \caption{Additional views of generated Manhattan blocks.}
    \Description{Additional views of generated Manhattan blocks.}
\end{figure*}

\begin{figure*}

     \begin{subfigure}{0.53\linewidth}
        \includegraphics[width=\linewidth]{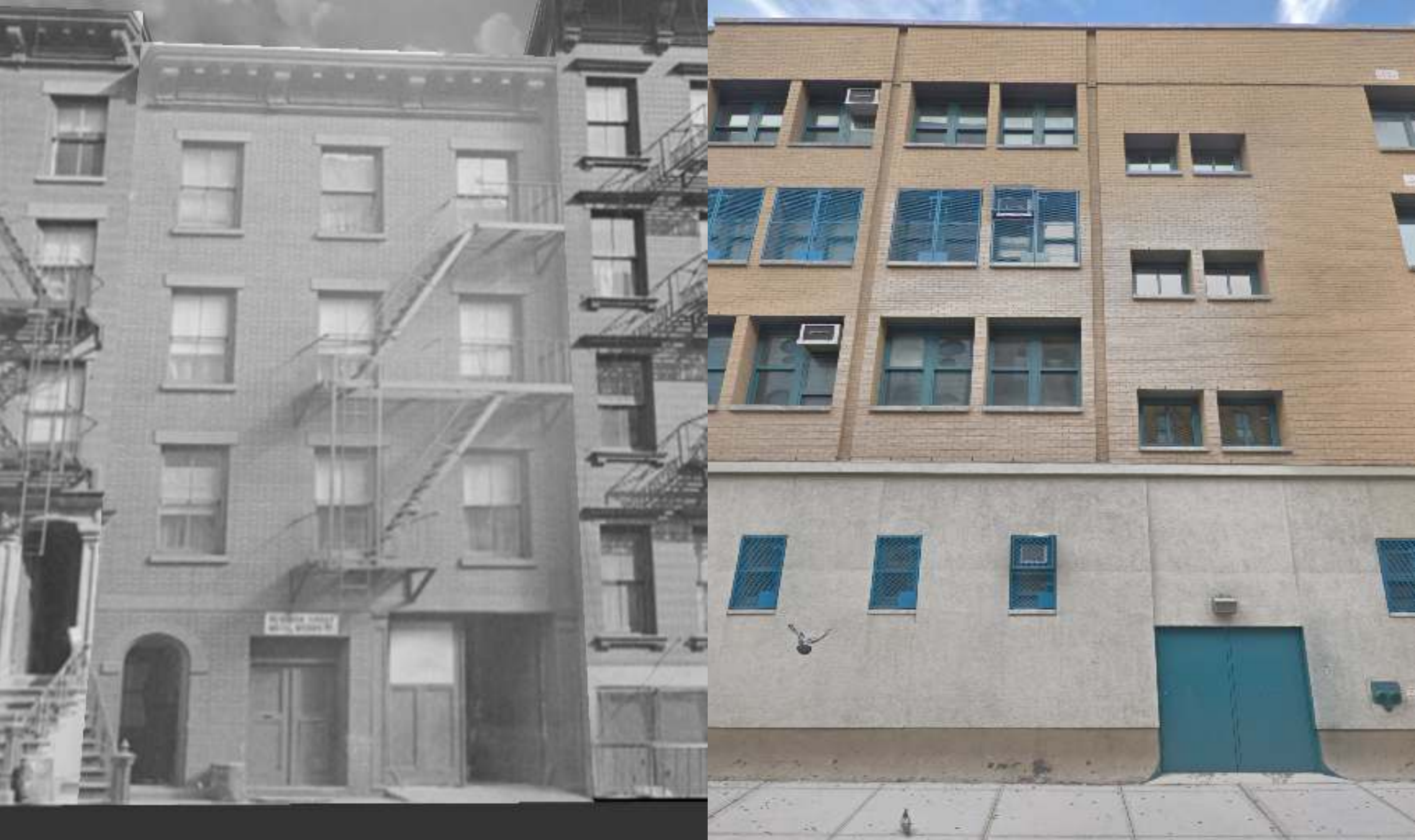}
        \caption{344, West 17th.}
    \end{subfigure}  
    \hfill
    \begin{subfigure}{0.45\linewidth}
        \includegraphics[width=\linewidth]{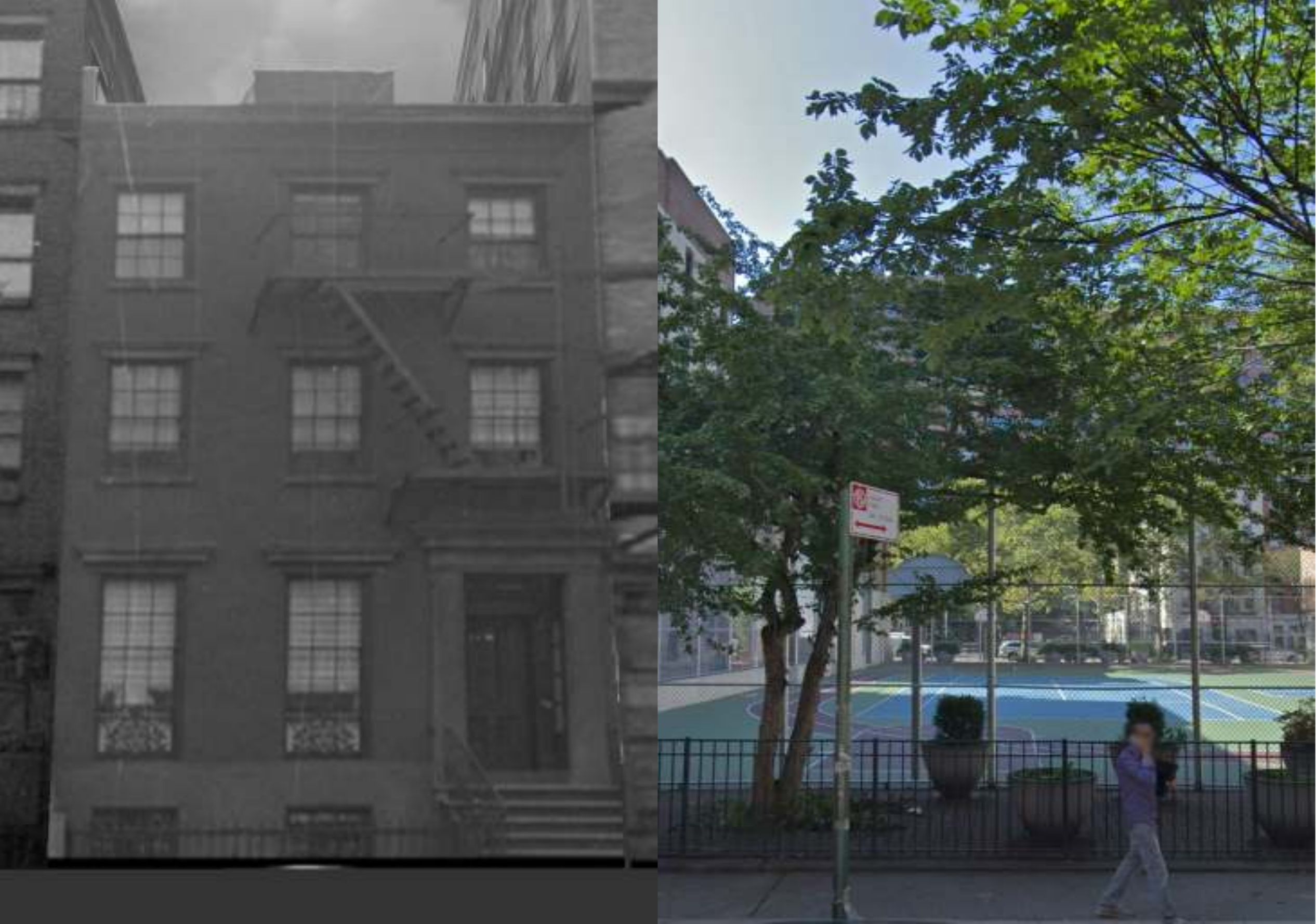}
        \caption{305, West 18th.}
    \end{subfigure}
    
    \begin{subfigure}{0.45\linewidth}
        \includegraphics[width=\linewidth]{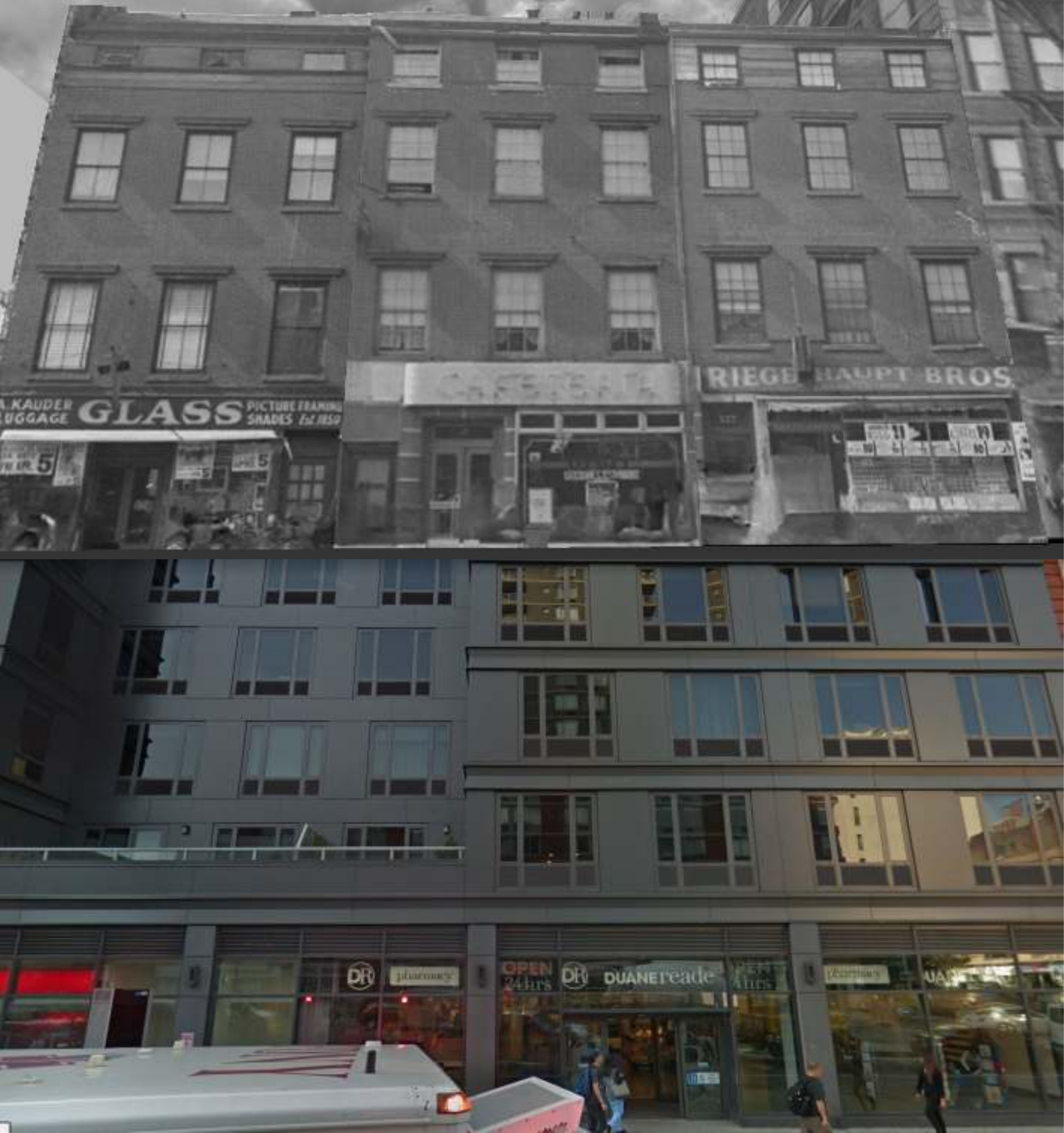}
        \caption{129, 8th Avenue.}
    \end{subfigure}
    \hfill
    \begin{minipage}{0.51\linewidth}
      \centering
        \begin{subfigure}{\linewidth}
            \includegraphics[width=\linewidth]{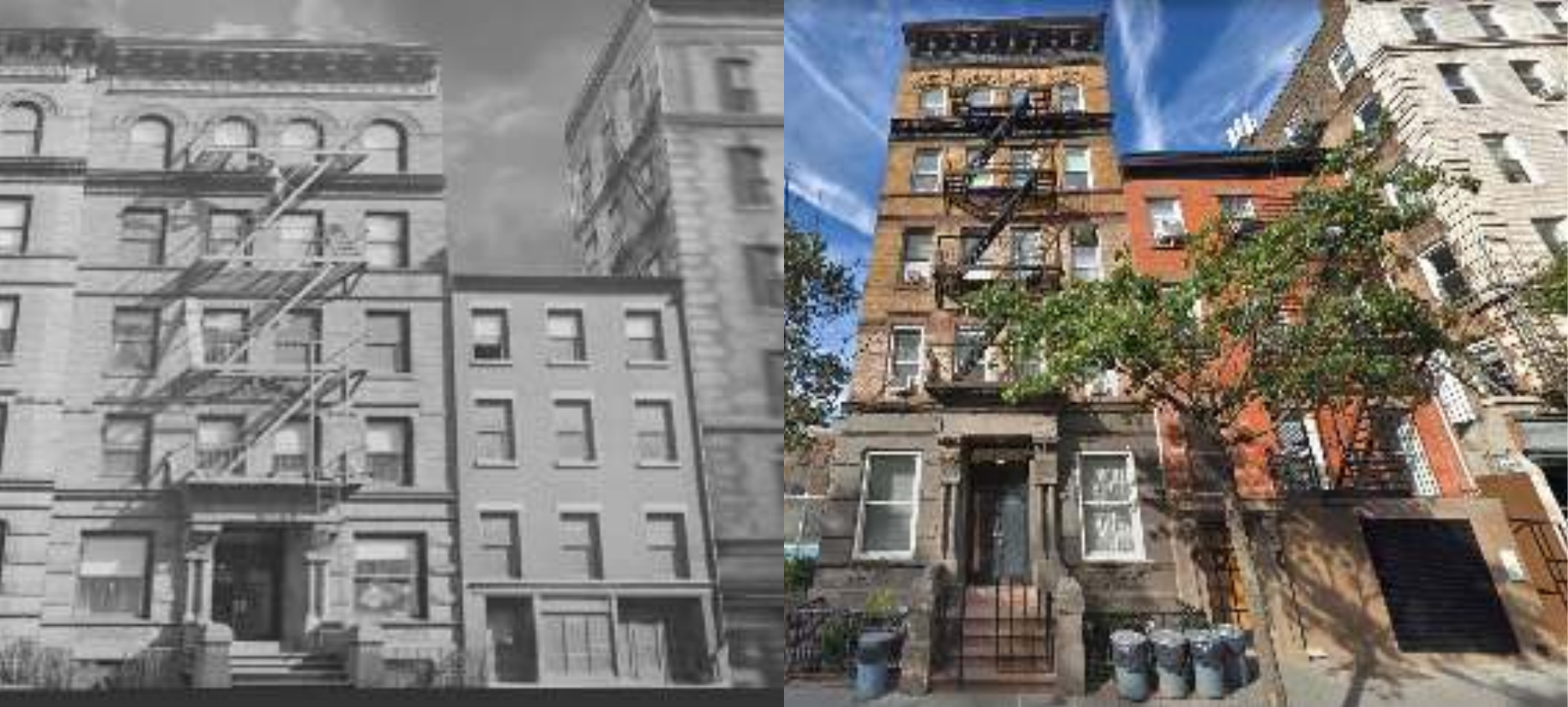}
            \caption{313, West 17th Street.}
        \end{subfigure}
        
        \begin{subfigure}{\linewidth}
            \includegraphics[width=\linewidth]{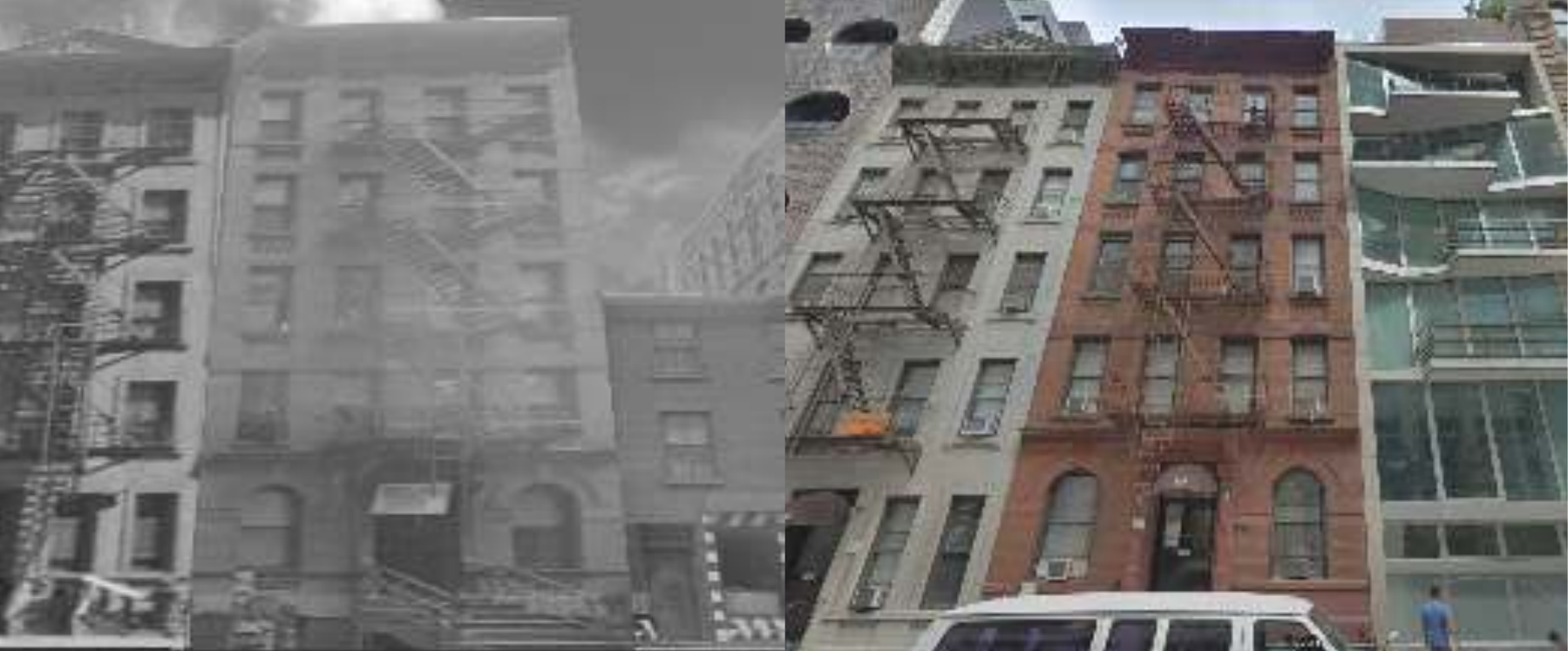}
            \caption{343, West 16th Street.}
        \end{subfigure}
    \end{minipage}
    
    \caption{Generated building models compared to modern day (Streetview).}
    \Description{Generated building models compared to modern day (Streetview).}
    \label{fig:more-3d}
\end{figure*}

\begin{figure*}
    \begin{subfigure}{0.32\textwidth}
        \includegraphics[width=\textwidth]{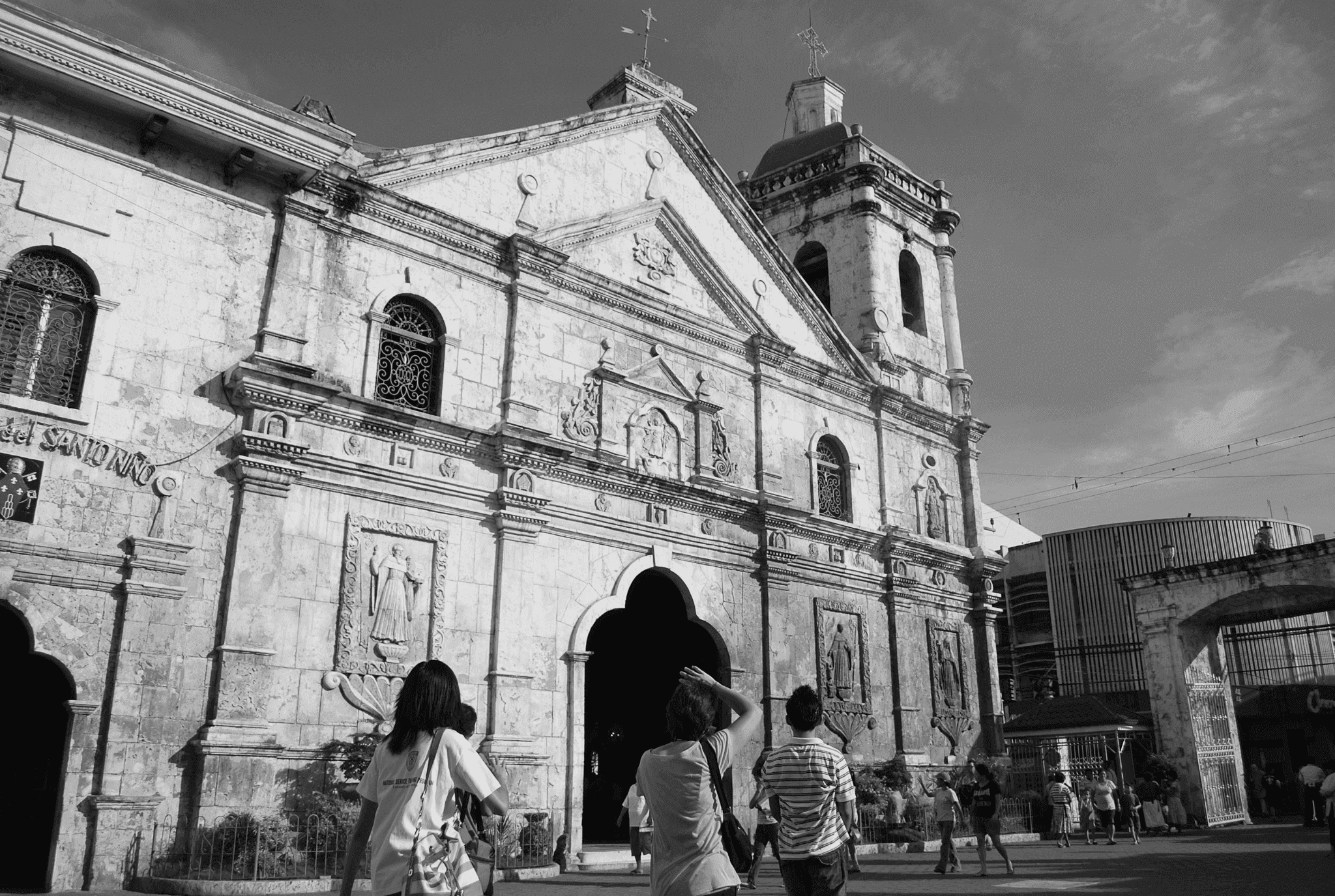}
        \caption{Input.}
    \end{subfigure}
    \begin{subfigure}{0.63\textwidth}
        \includegraphics[width=0.46875\textwidth]{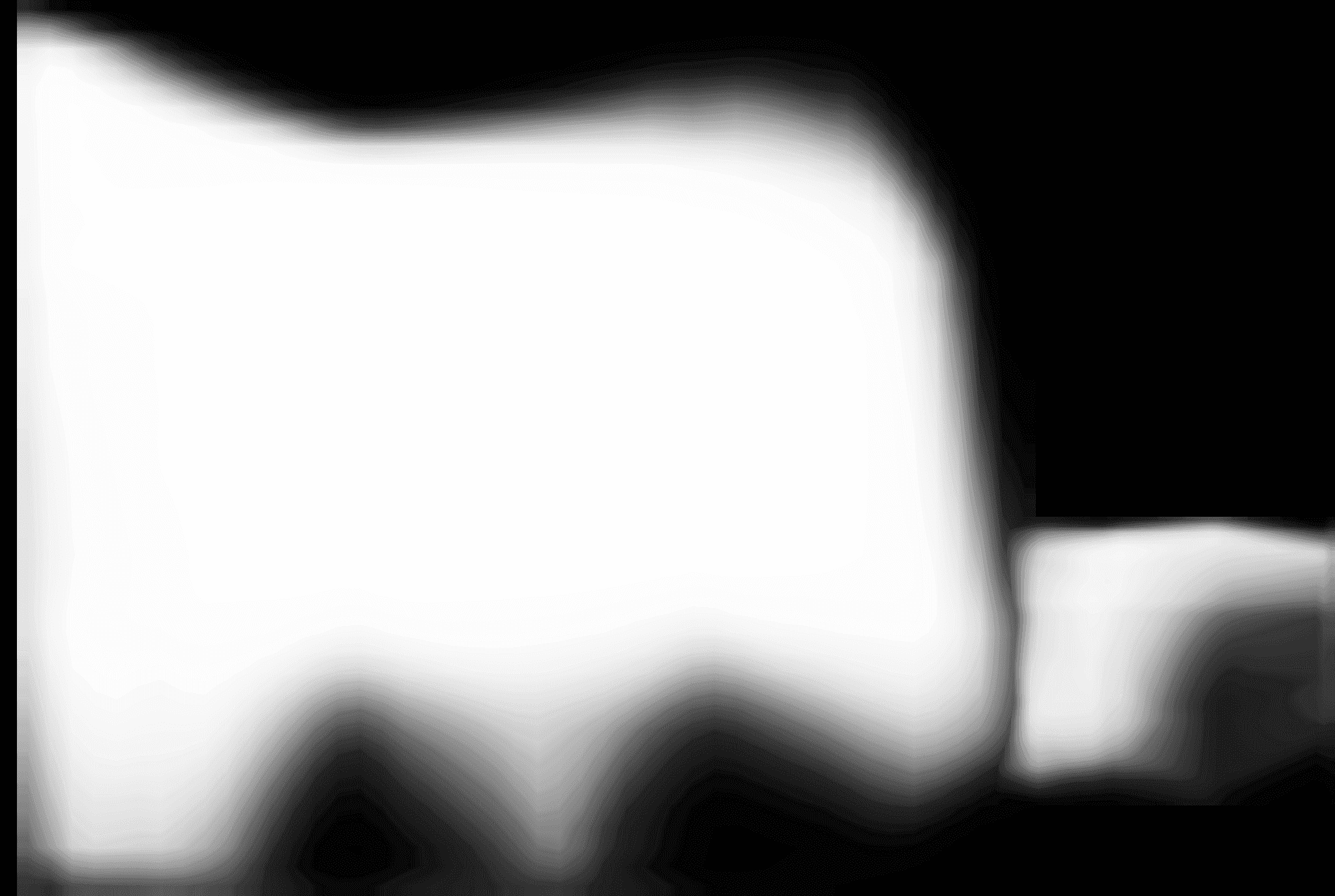}
        \includegraphics[width=0.46875\textwidth]{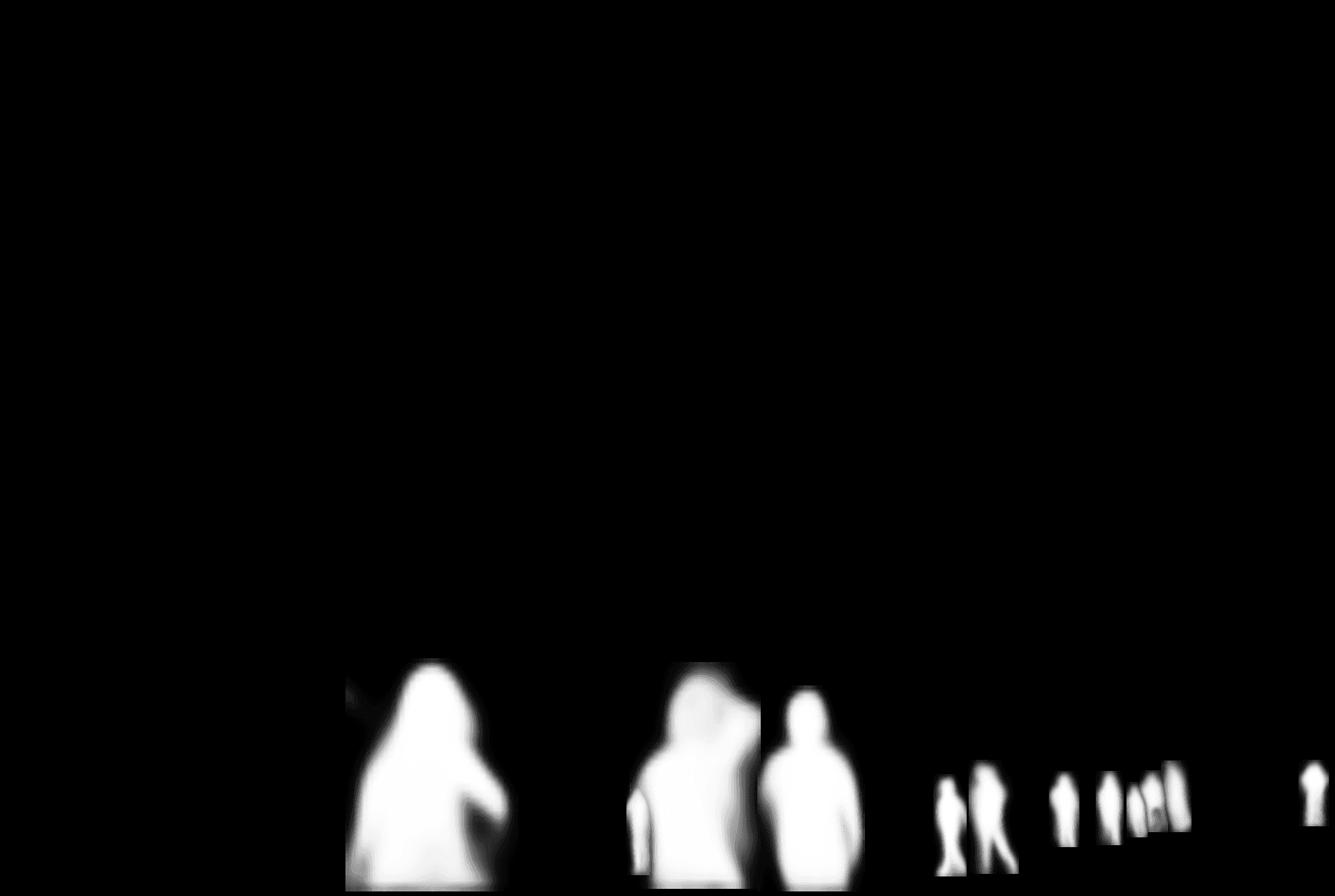}
        \caption{Segment Facade, Occlusions.}
    \end{subfigure}

    \begin{subfigure}{0.52\textwidth}
        \begin{minipage}{\linewidth}
                \includegraphics[width=0.46875\textwidth]{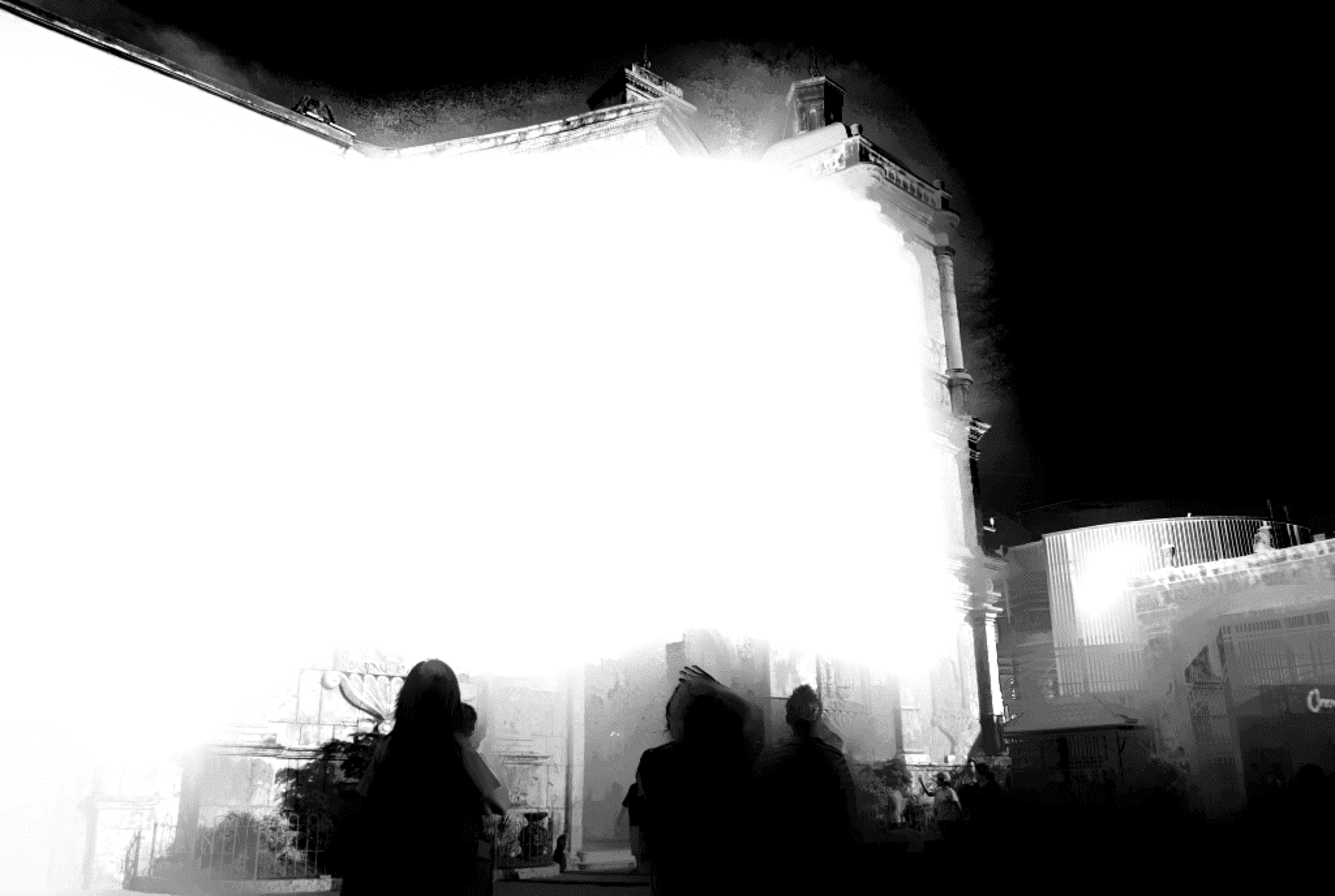}
                \includegraphics[width=0.46875\textwidth]{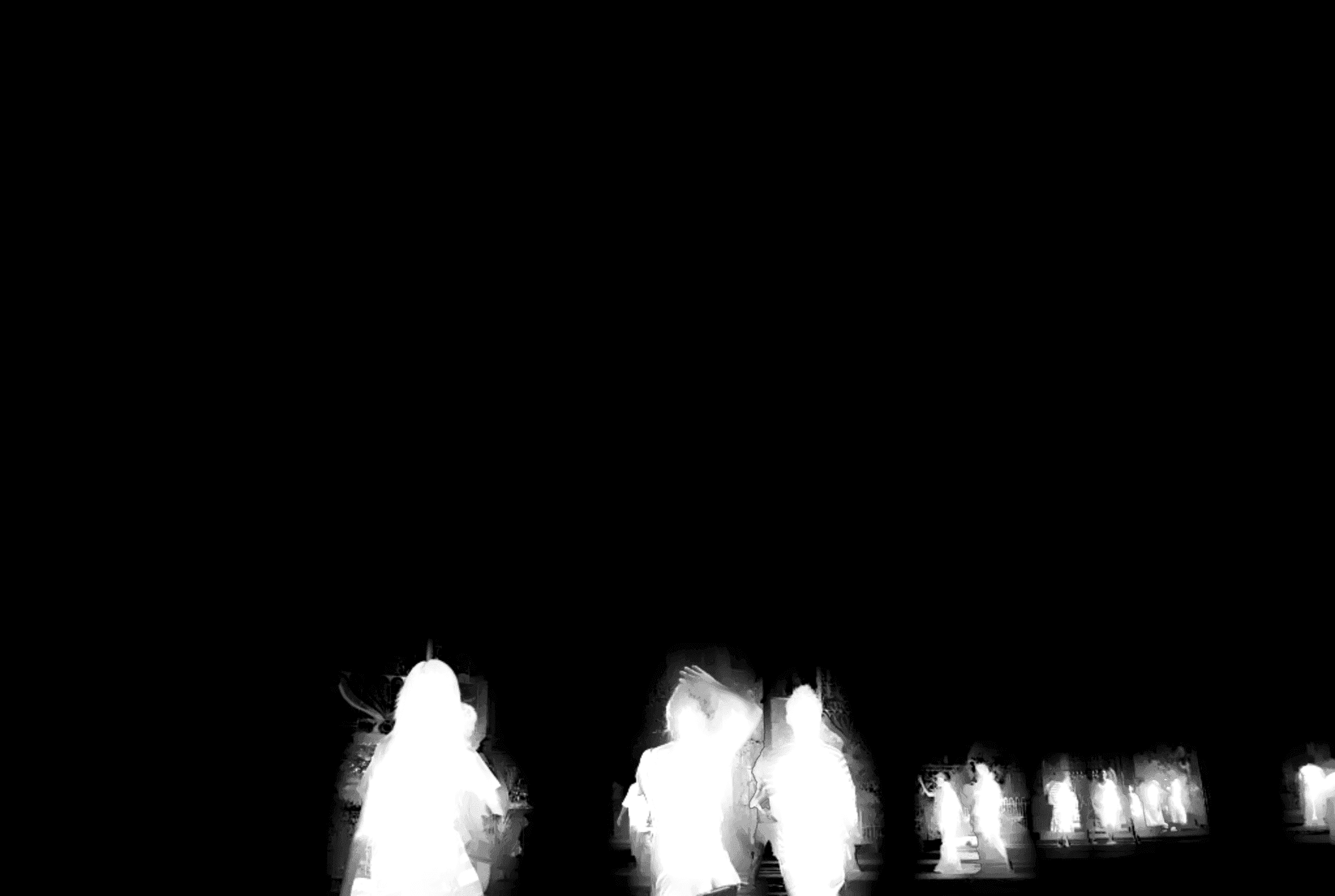}
                \caption{Mat Facade, Occlusions.}
                
                \includegraphics[width=0.46875\textwidth]{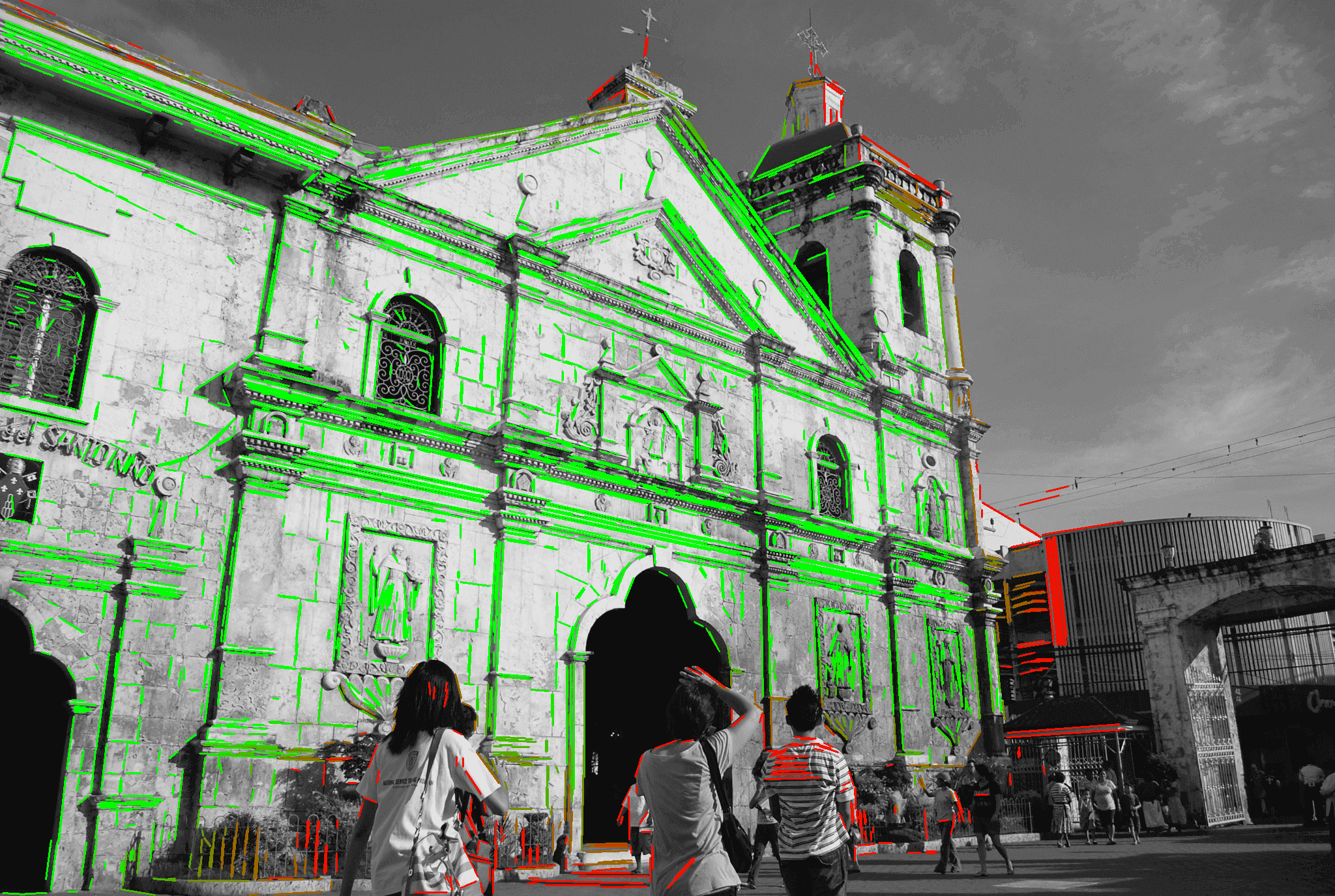}
                \includegraphics[width=0.46875\textwidth]{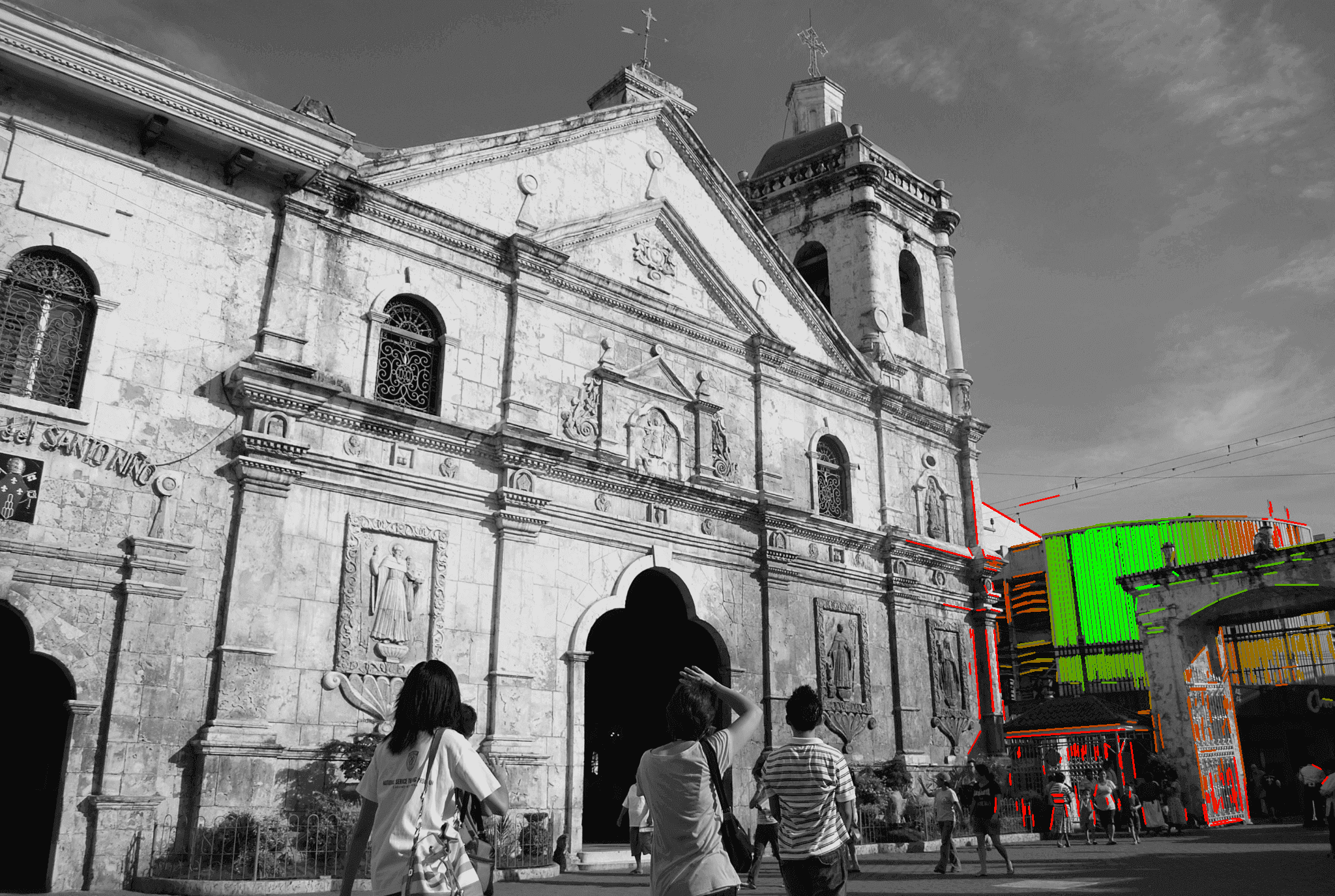}
                \caption{Extract Lines.}
        \end{minipage}
    \end{subfigure}
    \begin{subfigure}{0.43\textwidth}
        \includegraphics[width=0.46875\textwidth]{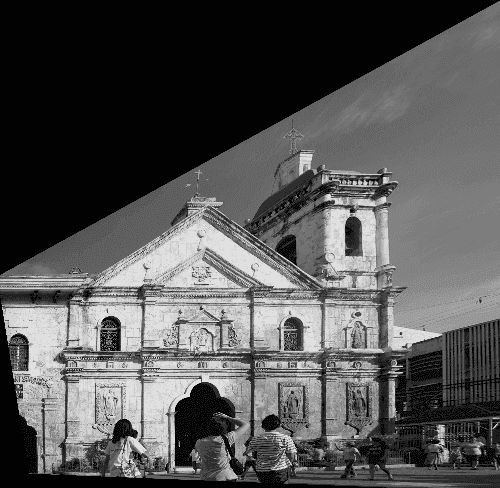}
        \includegraphics[width=0.46875\textwidth]{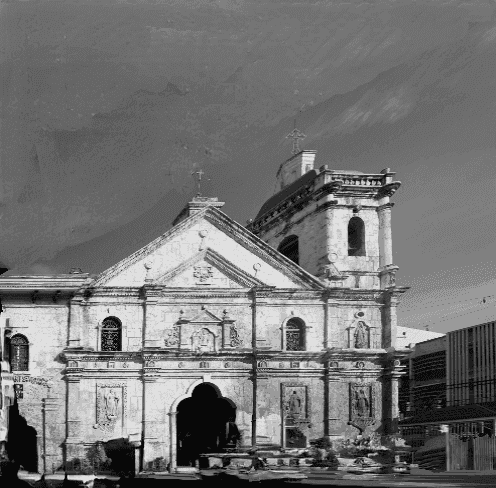}
            
        \includegraphics[width=0.46875\textwidth]{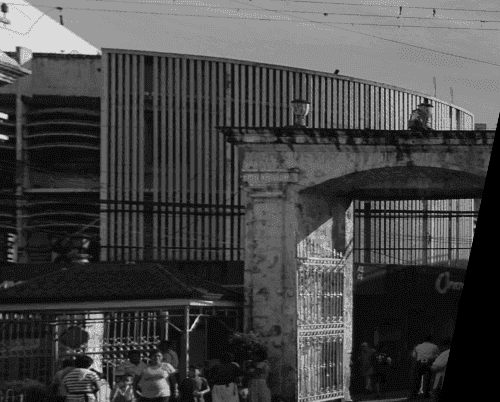}
        \includegraphics[width=0.46875\textwidth]{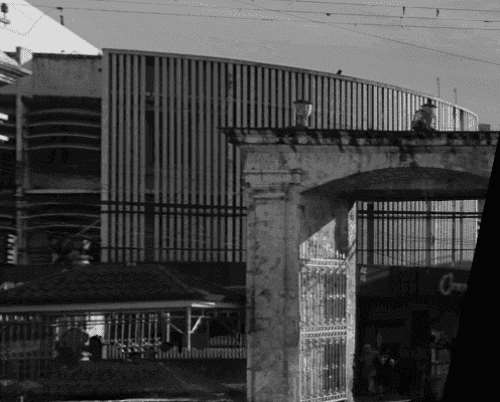}
        \caption{Rectification, Inpainting}
    \end{subfigure}

    \begin{subfigure}{0.5\textwidth}
        \centering
        \includegraphics[width=\textwidth]{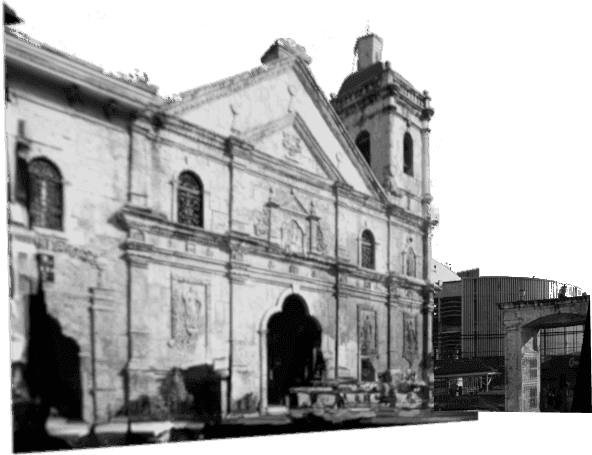}   
    \caption{Final Results}
    \end{subfigure}
    
	\caption{End to end processing pipeline depicting 2D facade extraction, rectification, and inpainting.}
	\Description{End to end processing pipeline depicting 2D facade extraction, rectification, and inpainting.}
\end{figure*}

\end{document}